%% file: main.tex
\documentclass[10pt]{article} 
\usepackage{microtype}
\usepackage{graphicx}
\usepackage{subcaption}
\usepackage{booktabs} 
\usepackage{amsmath}
\usepackage{xspace}
\usepackage{multirow}
\usepackage[dvipsnames]{xcolor}
\usepackage{amsmath}
\usepackage{amssymb}
\usepackage{mathtools}
\usepackage{amsthm}

\usepackage[accepted]{tmlr}

\input{math_commands.tex}

\usepackage{hyperref}
\usepackage[capitalize,noabbrev]{cleveref}
\usepackage{url}

\title{Enhancing Self-Supervised Visual Representation Learning via Low-Rank Adapted LLMs}

\author{\name Selim Kuzucu \email skuzucu@mpi-inf.mpg.de \\
      \addr Max Planck Institute for Informatics\\
      Saarland Informatics Campus
      \AND
      \name Muhammad Ferjad Naeem \email ferjad@google.com \\
      \addr Google
      \AND
      \name Anna Kukleva \email akukleva@mpi-inf.mpg.de\\
      \addr Max Planck Institute for Informatics \\
      Saarland Informatics Campus
      \AND
      \name Federico Tombari \email tombari@google.com\\
      \addr Google \\
      Technical University of Munich
      \AND
      \name Bernt Schiele \email schiele@mpi-inf.mpg.de\\
      \addr Max Planck Institute for Informatics \\
      Saarland Informatics Campus}

\def\month{08}  
\def\year{2026} 
\def\openreview{\url{https://openreview.net/forum?id=s2T8Kgj6Rd}} 

\begin{document}

\maketitle

\begin{abstract}
The integration of Large Language Model (LLM) blocks with Vision Transformers (ViTs) holds significant promise for vision-only tasks by leveraging the rich semantic knowledge and reasoning capabilities of LLMs. However, a fundamental challenge lies in the inherent modality mismatch between the text-centric pre-training of LLMs and the vision-centric training of ViTs. Direct fusion often fails to fully exploit the LLM's potential and suffers from unstable finetuning. Consequently, prior works typically keep LLM blocks frozen while learning only the vision components. To address these challenges, we introduce Language-Adapted Vision Enhancer (LAVIE), a novel framework that bridges this modality gap through a synergistic pre-training strategy. LAVIE co-adapts a ViT backbone and an LLM fusion block by (1) employing Masked Auto-Encoding (MAE) to pre-train the ViT for richer visual representations, and (2) concurrently training Low-Rank Adaptation (LoRA) layers within the LLM block using the same MAE objective. This joint optimization guides the ViT to produce LLM-aligned features and the LLM to effectively interpret visual information. We demonstrate through extensive experiments that LAVIE significantly improves performance in various downstream vision tasks, offering an effective and efficient way to enhance visual understanding using frozen LLM knowledge. Code is available at \href{https://www.github.com/selimkuzucu/LAVIE}{\texttt{ github.com/selimkuzucu/LAVIE}}.
\end{abstract}

\input{TMLR/Sections/1_Introduction}

\input{TMLR/Sections/2_Related_Work}
\input{TMLR/Sections/3_Methodology}
\input{TMLR/Sections/4_0_Experiments}
\input{TMLR/Sections/5_Analysis}
\input{TMLR/Sections/6_Conclusion}

\bibliography{main}
\bibliographystyle{tmlr}

\newpage
\appendix
\input{TMLR/Supplementary_Material/A_0_Appendix}
\end{document}

%% file: math_commands.tex
\usepackage{amsmath,amsfonts,bm}

\def\eqref#1{equation~\ref{#1}}

\def\1{\bm{1}}

\DeclareMathAlphabet{\mathsfit}{\encodingdefault}{\sfdefault}{m}{sl}
\SetMathAlphabet{\mathsfit}{bold}{\encodingdefault}{\sfdefault}{bx}{n}

\newcommand{\imp}[1]{\textcolor{ForestGreen}{+#1}}

\newcommand{\inimp}[1]{\textcolor{ForestGreen}{-#1}}
\newcommand{\ours}{LAVIE\xspace}
\newcommand{\myparagraph}[1]{\vspace{1.3pt}\noindent{\bf #1}}

%% file: TMLR/Sections/1_Introduction.tex
\section{Introduction}

The remarkable success of Large Language Models (LLMs)~\citep{brown2020language, touvron2023llama} has revolutionized natural language processing, demonstrating advanced capabilities in understanding, generation, and reasoning. This success has led to significant interest in extending their power to other modalities, particularly vision, impacting the field of Vision-Language Models (VLMs)~\citep{radford2021learning, alayrac2022flamingo, siglip2}. A promising direction within VLMs involves directly integrating powerful pre-trained LLM components with Vision Transformer (ViT)~\citep{dosovitskiy2020image} backbones, aiming to fuse visual models with the extensive semantic knowledge and  reasoning abilities learned by LLMs from vast textual corpora.

However, these applications of LLM for vision explore them in a generative framework, limiting their application to discriminative computer vision tasks. Pioneering works like LM4Vision~\citep{pang2023frozen} have explored fusing ViT features with terminal blocks of LLMs while learning a computer vision task, hinting at the potential benefits. Regardless, a critical hurdle persists: the alignment of representations originating from different modalities. LLMs are pre-trained exclusively on text, optimizing their internal representations for linguistic structures and concepts. Similarly, ViTs learn visual features optimized for tasks like image recognition. Simply injecting visual features into a text-centric LLM block often results in suboptimal alignment
\citep{liang2022mind}, where the LLM struggles to effectively ground its textual knowledge in the visual domain. Furthermore, adapting the large LLM component to the visual modality by joint fine-tuning can be computationally prohibitive and risks catastrophic forgetting or training instabilities~\citep{pang2023frozen, lai2024residual}.
\begin{figure*}[ht]

  \centering
  \scalebox{0.85}{\includegraphics[width=\linewidth]{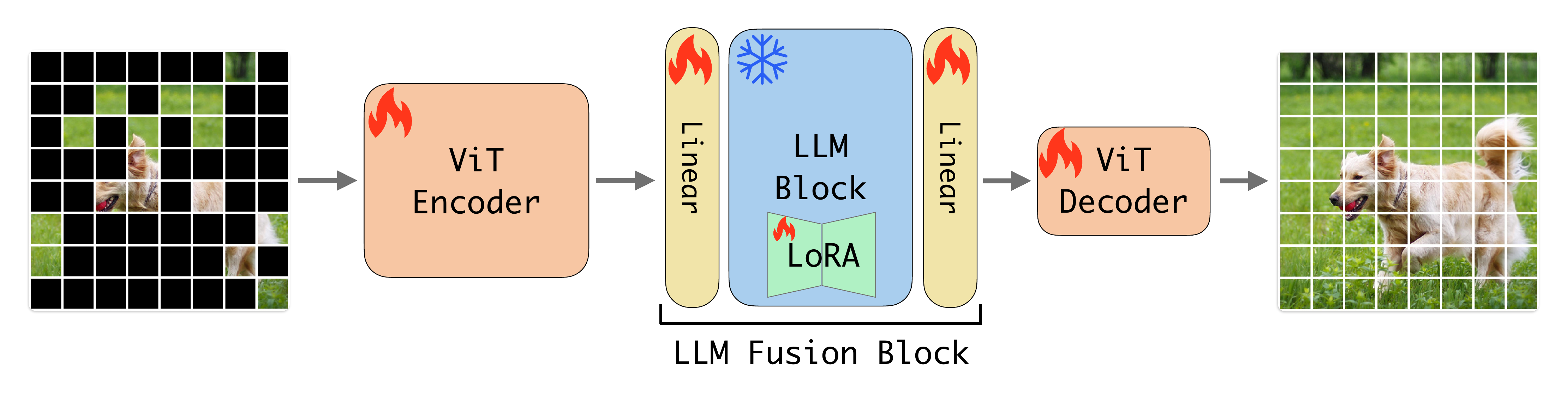}}

  \caption{Architecture diagram of our \textbf{Language-Adapted Vision Enhancer (\ours)}. Input image patches are processed by the ViT Encoder. The resulting visual features are then passed through an LLM Fusion Block (comprising linear projections and an LLM transformer block adapted with LoRA). For MAE pre-training, a lightweight decoder reconstructs masked patches. For fine-tuning, the decoder is removed, and a task-specific head is added.}
  \label{fig:teaser_mae}

\end{figure*}

To address these challenges, we introduce \textbf{L}anguage-\textbf{A}dapted \textbf{VI}sion \textbf{E}nhancer (\ours), a novel framework designed to foster a more profound and efficient synergy between ViTs and LLMs for discriminative vision tasks. Our core idea is a two-fold strategy:
\begin{enumerate}
    \item \textbf{Enhanced Visual Representation Learning:} We pre-train the ViT backbone using Masked Auto-Encoding (MAE)~\citep{he2022masked}. This self-supervised objective encourages the ViT to learn richer, more context-aware visual representations that we hypothesize are more informative for the LLM blocks.
    \item \textbf{Efficient LLM Adaptation and Modality Bridging:} Simultaneously, we adapt the fused LLM block (e.g., from LLaMA) using Low-Rank Adaptation (LoRA)~\citep{hu2022lora}. Crucially, these LoRA layers are trained \textit{concurrently} with the MAE pre-training of the ViT, using the same MAE reconstruction loss. This joint optimization allows the LLM to efficiently learn to interpret the evolving visual features, effectively translating its vast semantic knowledge to the visual domain without requiring full fine-tuning of the LLM.
\end{enumerate}
This synergistic pre-training process is key: the ViT learns to leverage the pretrained LLM representations, while the LLM (via LoRA) learns to ``understand'' these visual features, thereby bridging the modality mismatch from both ends. Our contributions are fourfold:

\begin{itemize}
    \item We propose \ours, a novel architecture and pre-training strategy that co-adapts a ViT and an LLM block through joint MAE-based self-supervision and LoRA-based LLM adaptation, effectively mitigating the alignment issue between representations of different modalities.
    \item We demonstrate that this concurrent optimization of LoRA layers within the LLM during MAE pre-training enables efficient and stable adaptation of the LLM, allowing it to effectively leverage its textual knowledge for visual understanding.
    \item We show through extensive experiments on benchmark computer vision tasks that \ours significantly outperforms existing LLM-fusion approaches that employ more direct strategies, pushing forward leveraging frozen LLM capabilities in vision models.
    \item We provide intriguing analyses regarding the attention entropies of \ours and how it achieves stronger performance through improved background robustness.
\end{itemize}

%% file: TMLR/Sections/2_Related_Work.tex
\section{Background and Related Work}
\myparagraph{Self-supervised learning.} Self-supervised learning (SSL) has emerged as a powerful paradigm for leveraging readily available unlabeled data.
SSL methods have achieved widespread success in the broader machine learning community, starting with earlier contrastive approaches \citep{chen2020simple, he2020momentum}, achieving new frontiers in representation learning otherwise unreachable with full-supervised techniques.
More recently, SSL approaches have powered foundation models in a wide range of domains, from NLP \citep{touvron2023llama, touvron2023llama2, devlin2019bert} to vision \citep{caron2021emerging, grill2020bootstrap, naeem2024silc}.
\myparagraph{Masked image modeling.} Masked image modeling is an established example of self-supervised learning methods for computer vision, initially pioneered by stacked denoising autoencoders \citep{vincent2010stacked}.
Motivated by the success of masked language modeling approach of BERT \citep{devlin2019bert}, a plethora of follow-up works proposed novel self-supervised masked image modeling techniques \citep{chen2020generative, bao2021beit, zhou2021ibot, dosovitskiy2020image}.
Among these works, Masked Auto-Encoders (MAE) \citep{he2022masked} stand out with their accelerated pretraining approach consisting of a heavyweight encoder observing only a small fraction of image patches and a lightweight decoder reconstructing the original image features.
MAE has established itself as a strong approach not only for global image recognition but also for more challenging fine-grained visual recognition tasks, such as object detection \citep{li2022exploring}.

{\myparagraph{Multimodal vision-language models.} The integration of vision and language has largely been driven by Multimodal Large Language Models (MLLMs), which frequently employ parameter-efficient fine-tuning to align pre-trained visual encoders with language decoders \citep{li2023blip, liu2023visual, alayrac2022flamingo}.
These architectures are predominantly designed for generative tasks, such as visual question answering or image captioning \citep{goyal2017making, chen2015microsoft}, and inherently rely on cross-modal training objectives and textual data.
While highly effective for language generation, these frameworks are fundamentally distinct from our setting.
\ours operates entirely within a unimodal, self-supervised representation learning paradigm.
Rather than adapting a pre-existing vision-language pipeline for downstream text generation, \ours leverages the semantic priors of a pretrained LLM block to enhance the foundational pre-training of the vision transformer itself, targeting discriminative computer vision tasks without requiring any text inputs or generative decoding.
We provide a more comprehensive discussion with these works in Appendix \ref{app-sec:comparison-with-lvlms}.

\myparagraph{Using frozen LLM blocks for visual tasks.} Close to our work are the works that directly employ frozen pretrained LLM blocks with vision transformers \citep{pang2023frozen, lai2024residual, bai2025frozen}.
Among these, \citet{pang2023frozen} is the pioneering work that showed that using frozen LLM blocks on top of vision transformers can provide strong performance gains on pure vision tasks.
However, \citet{pang2023frozen} did not aim to achieve \textit{competitive} performance on visual recognition but rather provided relative performance improvement on a wide range of vision tasks.

In this work, we combine the powers of self-supervised learning, initial explorations of \citet{pang2023frozen}, and LoRA adaptations together to achieve significantly improved downstream performance, differing from the previous art.
Evidenced by our experiments, our work provides stronger recipes for achieving robust visual recognition performance while better leveraging the LLM blocks.

%% file: TMLR/Sections/3_Methodology.tex
\section{\ours: Language-Adapted Vision Enhancer}
\label{sec:methodology}
While LM4Vision~\citep{pang2023frozen} demonstrated the potential of fusing Vision Transformers (ViTs) with the terminal block of a Large Language Model (LLM), the direct introduction of this transformer block introduces a modality mismatch due to the LLM's text-centric pre-training and Vision Transformer's visual processing. To address this, we propose a two-fold strategy. First, we introduce Self-Supervised Learning (SSL) using Masked Auto-Encoding (MAE) during the pre-training of the ViT backbone. This step aims to better align visual representations with the language modality. Second, to adapt the LLM component (e.g., LLaMA), pre-trained solely on text, we incorporate Low-Rank Adaptation (LoRA). This allows the LLM to efficiently translate its extensive semantic knowledge, learned from billion-scale textual data, to the visual domain, thereby improving performance on target computer vision tasks.

\subsection{\ours: Language-Adapted Vision Enhancer}
We introduce \textbf{L}anguage-\textbf{A}dapted \textbf{VI}sion Enhancer (\ours), with the aim of effectively bridging the representation alignment issue between vision and language representations when using language trained transformer blocks in vision transformers. The core intuition is to enable a synergistic co-adaptation: the ViT learns to produce visual features amenable to language processing, while the LLM block learns to interpret these visual features, all within a unified pre-training framework.

Our \ours architecture (illustrated in Figure~\ref{fig:teaser_mae}) comprises of three main components:
\begin{enumerate}
    \item \textbf{Vision Transformer (ViT) Encoder ($\mathbf{M}_{Enc}$):} Following \citep{dosovitskiy2020image}, the standard ViT maps input patches $x$ into latent visual representations $z_v=\mathbf{M}_{Enc}(x)$.
    \item \textbf{LLM Fusion Block ($\mathbf{M}_{LLM}^{\text{fuse}}$):} This module integrates a pre-trained LLM transformer block (e.g., from LLaMA~\citep{touvron2023llama}) into the pipeline to enrich the visual features $z_v$. To manage differing hidden dimensions and facilitate adaptation, $z_v$ is first projected by a linear layer $\mathbf{M}^1_{L}$, then processed by the LLM block $\mathbf{M}_{LLM}$, and finally projected back by $\mathbf{M}^2_{L}$. Thus, the enhanced latent features are $z'_{v} = \mathbf{M}^2_{L} \cdot \mathbf{M}_{LLM} \cdot \mathbf{M}^1_{L}(z_v)$. We denote this entire compound mapping as $\mathbf{M}_{LLM}^{\text{fuse}}(z_v) \rightarrow z'_v$.
    \item \textbf{Lightweight MAE Decoder ($\mathbf{M}_{Dec}$):} For self-supervised pre-training, a shallow transformer decoder, similar to \citep{he2022masked}, takes the enhanced latent features $z'_v$ from visible patches and reconstructs the original masked image patches $x'$.
\end{enumerate}

The complete pre-training pipeline for an input image $x$ can thus be expressed as:
\begin{equation}
    x' = \mathbf{M}_{Dec} \left( \mathbf{M}_{LLM}^{\text{fuse}} \left( \mathbf{M}_{Enc}(x_{\text{vis}}) \right), x_{\text{mask\_ids}} \right),
    \label{eq:alvit-pipeline}
\end{equation}
where $x_{\text{vis}}$ represents  visible patches fed to the encoder, and $x_{\text{mask\_ids}}$ represents information about the masked patches required by the decoder for reconstruction (e.g., their positional encodings).

\subsection{Synergistic Pre-training for Modality Alignment}
The core component of \ours is its pre-training strategy, designed to address the modality mismatch through self-supervised pretraining. This involves concurrently training the ViT via Masked Auto-Encoding (MAE) and adapting the LLM fusion block using LoRA.

\subsubsection{Self-Supervised Visual Representation Learning via MAE}
\myparagraph{Intuition.} Standard ViT training (e.g., on ImageNet) learns features optimized for classification but these features often fail to capture deeper semantics required for other computer vision tasks~\citep{he2022masked}. However, self-supervised pretrained backbones learn more generic features often directly usable across a plethora of computer vision tasks~\citep{oquab2023dinov2, he2022masked}. We utilize Masked Auto-Encoding~(MAE)~\citep{he2022masked} as the self-supervision framework owing to its recent success in learning robust features and its efficiency~\citep{he2022masked, siglip2, li2022exploring}. MAE learns holistic and context-aware representations by reconstructing heavily masked inputs. When learned together with a LLM block, we hypothesize that such representations are inherently richer and more compatible with the high-level understanding capabilities of LLM block.

\myparagraph{Mechanism.} We follow the standard MAE pre-training strategy proposed by ~\citep{he2022masked}. An input image $x$ is divided into $N$ non-overlapping patches. A high percentage (e.g., 75\%) of these patches are randomly masked out. Only the visible patches $x_{\text{vis}}$ are processed by the ViT encoder $\mathbf{M}_{Enc}$ and subsequently by the LLM fusion block $\mathbf{M}_{LLM}^{\text{fuse}}$. The lightweight decoder $\mathbf{M}_{Dec}$ takes the output from the LLM block and reconstructs the original pixels of the masked patches from the enhanced latent representations $z'_v$ and the positional embeddings of all patches. The learning objective minimizes the Mean Squared Error (MSE) between the reconstructed and original masked patches. This process trains the ViT backbone $\mathbf{M}_{Enc}$.

\subsubsection{Efficient LLM Adaptation with Low-Rank Adaptation (LoRA)}
\myparagraph{Intuition.} Pre-trained LLMs possess vast world knowledge and complex reasoning abilities encoded in their weights. Fine-tuning the entire LLM for a vision task is computationally prohibitive and risks catastrophic forgetting of its semantic understanding capabilities that we want to utilize for visual understanding. LoRA~\citep{hu2022lora} offers a parameter-efficient solution, allowing us to "steer" the LLM's knowledge towards the visual domain by training only a small number of additional parameters. It also allows for stable finetuning of the LLM block without the risk of the larger LLM block collapsing the training signal.

\myparagraph{Mechanism.} We inject LoRA layers into the query ($W_q$) and value ($W_v$) projection matrices of the LLM block $\mathbf{M}_{LLM}$. For a pre-trained weight matrix $W_0 \in \mathbb{R}^{d \times k}$, its update is represented by a low-rank decomposition $W_0 + \Delta W = W_0 + BA$, where $B \in \mathbb{R}^{d \times r}$, $A \in \mathbb{R}^{r \times k}$, and the rank $r \ll \min(d, k)$. Only $A$ and $B$ are trainable. The original LLM weights $W_0$ remain frozen keeping their pre-trained knowledge secure.

\subsubsection{Joint Optimization: The Key to Modality Bridging}
A critical aspect of our method is that the LoRA layers within $\mathbf{M}_{LLM}^{\text{fuse}}$ are trained \textit{concurrently} with the ViT backbone during the MAE pre-training phase. The MAE reconstruction loss not only guides the ViT but also backpropagates through the LLM fusion block, updating the LoRA parameters.
This joint optimization fosters a synergistic co-adaptation, while the  ViT ($\mathbf{M}_{Enc}$) learns to produce visual embeddings that are not only good for reconstruction but also effectively processed and enhanced by the LLM block. The LLM block
learns to interpret and refine these evolving visual embeddings via LoRA in $\mathbf{M}_{LLM}$, leveraging its pre-trained frozen textual knowledge to enhance them with richer semantics relevant to the visual context.
This simultaneous learning process is crucial for bridging the modality mismatch, as it forces the two modalities to be jointly aligned rather than adapting one to a fixed representation of the other. The LLM is not just passively processing ViT features; it is actively being aligned to understand the visual domain while the ViT learns to present this information in a more digestible format in the LLM space.

\subsection{Architectural Adjustments for Cross-Modal LLM Processing}
To further enhance the LLM block's suitability for processing visual information, we incorporate specific architectural modifications, following the existing works~\citep{pang2023frozen, lai2024residual}.
\textbf{(1) Bidirectional Attention.} LLMs commonly use causal attention masks. However, visual information in an image does not possess sequential causality the same way as language. Thus, we replace the causal attention mechanism in the LLM block with bidirectional attention. This allows each visual token representation in the LLM block to attend to all others, allowing for a holistic understanding.
\textbf{(2) Removal of Rotary Pos. Embeddings (RoPE).} RoPE~\citep{su2024roformer}, commonly used in LLMs, encodes absolute and relative positional information tailored for text sequences. Since our ViT backbone already incorporates learned positional embeddings for visual patches, and the nature of spatial relationships in images differs from sequential text, we remove RoPE from the LLM block. This simplifies the architecture, prevents the imposition of text-specific biases onto visual features, and ensures consistency with typical ViTs. 

\subsection{Downstream Fine-tuning}
After the MAE-based pre-training with joint LoRA adaptation, \ours is fine-tuned for specific downstream computer vision tasks (e.g., image classification). For fine-tuning, we discard the MAE decoder ($\mathbf{M}_{Dec}$), and add a task-specific head (e.g., a linear classifier) on top of the output features $z'_v$. 
During fine-tuning, the ViT backbone, the linear projection layers $\mathbf{M}^1_L, \mathbf{M}^2_L$, and the LoRA parameters within the LLM block can be further trained. The original weights of the LLM block $\mathbf{M}_{LLM}$ remain frozen, preserving its extensive learned knowledge while allowing targeted adaptation through LoRA. This strategy ensures efficient transfer of learned representations to downstream tasks.

%% file: TMLR/Sections/4_0_Experiments.tex
\section{Experiments}
\label{sec:experiments}
\input{TMLR/Sections/4_1_Implementation_Details}
\input{TMLR/Sections/4_2_Experiments_Classification}

\input{TMLR/Sections/4_3_Ablations}

%% file: TMLR/Sections/4_1_Implementation_Details.tex
We now discuss our experiments and highlight the strengths of our Language-Adapted Vision Enhancer (\ours).

\myparagraph{Datasets.} For our image classification experiments, we utilize the ImageNet-1K benchmark \citep{deng2009imagenet}.
In addition, we report evaluation results on several domain-shift benchmarks, namely ImageNet-C \citep{hendrycks2019benchmarking}, ImageNet-A \citep{hendrycks2021natural}, ImageNet-SK \citep{wang2019learning}, ImageNet-V2 \citep{recht2019imagenet}, ImageNet-R \citep{hendrycks2021many} and Imagenet-E \citep{li2023imagenete}.
We also report results on ImageNet-9 benchmark \citep{xiao2020noise}, which measures the reliance of a model on background and foreground features.
Among its splits, we choose the \textit{mixed same} and the \textit{mixed random}.
In the former, backgrounds of images are randomly replaced with the background of another image of the same class, and in the latter the background is replaced with the background of an image of a completely random class.
For fine-grained visual recognition, we utilize MS COCO~\citep{lin2014mscoco} for object detection, DAVIS-2017~\citep{pont20172017} for video object segmentation, and ImageNet-Segmentation~\citep{gao2022large}.
We adhere to the evaluation protocols of \citet{li2022exploring} for MS COCO, \citet{caron2021emerging} for DAVIS-2017, and \citet{pang2023frozen} for ImageNet-Segmentation.

\myparagraph{Pretraining.} Our pre-training settings closely mirror that of the original MAE work \cite{he2022masked}, including all of the hyperparameters related to the training (learning rate, mask ratio, \textit{etc.}).
We pre-train both vanilla MAE-ViT baselines and our \ours for a total of $800$ epochs, following \cite{he2022masked}.
For our LLM block, unless otherwise specified, we utilize the $32^{nd}$ transformer block of LLaMA 1 \cite{touvron2023llama}, following our ablations in Section \ref{sec:ablation}.
As described in Section \ref{sec:methodology}, while the original LLM weights are always kept frozen, we also integrate LoRA \citep{hu2022lora} into the $Q$ and $V$ projection matrices, both with a rank of $16$, constituting a very small fraction ($0.3\%$) of the number of trainable parameters. At inference time, this module remains active and adds a single frozen LLM transformer block to the forward pass, increasing forward-pass compute relative to a vanilla ViT.

\myparagraph{End-to-end Finetuning.} For image classification, we perform finetuning for $100$ epochs on both the baselines and our \ours after the pre-training stage, while adhering to all of the hyperparameter settings and other training details presented in \citep{he2022masked}.
For fine-grained visual recognition, we train both the baselines and \ours for $100$ epochs after pre-training, while adhering to all training settings in ViTDet \citep{li2022exploring}.
Finally, with the exception of Tables \ref{tab:imagenet-main-results} \& \ref{tab:primary-ablations}, all results are reported with a random seed of $0$, following our baselines.

%% file: TMLR/Sections/4_2_Experiments_Classification.tex
\subsection{Image Classification}
\label{sec:image-classification-exp}
\begin{table*}[t]
\centering
\caption{\ours achieves significantly better Top-1 accuracy (\%) in frozen LLM augmented model setting on IN-1K, drastically improving over the supervised baselines. We also demonstrate significantly enhanced robustness across challenging variants (IN-A, IN-SK, IN-V2, IN-R, IN-C, IN-E). \ours consistently outperforms both supervised baselines and the strong MAE-pretrained ViT/B. Each result denotes the average of $3$ random seeds along with associated standard deviations. To obtain the standard deviations, we reproduced the results of \citet{pang2023frozen}, and provide the original numbers in \textcolor{gray}{gray} for reference. \textbf{Bold} indicates the best result.}
\label{tab:imagenet-main-results}

\scalebox{0.78}{
\begin{tabular}{c|c||c|c|c|c|c|c|c}
\toprule
\midrule
Training & Model & IN-1K & IN-A & IN-SK & IN-V2 & IN-R & IN-C & IN-E \\
\midrule
Supervised-Only & \textcolor{gray}{ViT/B*} &
\textcolor{gray}{$80.6$} & \textcolor{gray}{$23.4$} & \textcolor{gray}{$31.9$} &
\textcolor{gray}{$-$} & \textcolor{gray}{$43.5$} & \textcolor{gray}{$60.2$} & \textcolor{gray}{$-$}\\
{[LM4Vision]} & \textcolor{gray}{ViT/B+LM1*} &
\textcolor{gray}{$81.7$} & \textcolor{gray}{$26.9$} & \textcolor{gray}{$33.2$} &
\textcolor{gray}{$-$} & \textcolor{gray}{$44.3$} & \textcolor{gray}{$62.1$} & \textcolor{gray}{$-$}\\
\citep{pang2023frozen} & ViT/B &
$79.58_{\pm0.81}$ & $22.78_{\pm3.77}$ & $30.61_{\pm0.68}$ &
$67.48_{\pm1.07}$ & $42.57_{\pm1.42}$ & $59.73_{\pm1.69}$ & $80.17_{\pm2.15}$\\
& ViT/B+LM1 &
$80.50_{\pm0.25}$ & $23.22_{\pm0.80}$ & $31.06_{\pm0.48}$ &
$68.69_{\pm0.41}$ & $41.92_{\pm0.57}$ & $61.24_{\pm0.27}$ & $80.83_{\pm0.12}$\\
\midrule
MAE Pretrained & ViT/B &
$83.11_{\pm0.09}$ & $33.64_{\pm0.11}$ & $35.69_{\pm0.30}$ &
$72.73_{\pm0.21}$ & $49.88_{\pm0.32}$ & $62.86_{\pm0.01}$ & $86.43_{\pm0.06}$ \\
& \ours/B \textit{(Ours)} &
$\mathbf{83.63_{\pm0.04}}$ & $\mathbf{36.39_{\pm0.28}}$ & $\mathbf{36.36_{\pm0.61}}$ &
$\mathbf{73.15_{\pm0.02}}$ & $\mathbf{50.17_{\pm0.16}}$ & $\mathbf{63.44_{\pm0.05}}$ & $\mathbf{86.97_{\pm0.15}}$\\
& &
$\imp{0.52}$ & $\imp{2.75}$ & $\imp{0.67}$ & $\imp{0.42}$ & $\imp{0.29}$ & $\imp{0.58}$ & $\imp{0.54}$\\
\midrule
\bottomrule
\end{tabular}
}
\end{table*}

We evaluate \ours on ImageNet-1K and its variants designed to test robustness to domain shifts (IN-A, IN-SK, IN-V2, IN-R) and common corruptions (IN-C). Furthermore, we replicated the results of \citet{pang2023frozen} using the authors' code to obtain the standard deviation, as they were reported only with a single seed.
Results in Table~\ref{tab:imagenet-main-results} demonstrate the performance improvements of our \ours.

\myparagraph{\ours Outperforms All Baselines.}
Our \ours/B model significantly outperforms previous works, achieving $\mathbf{83.63}\%$ top-1 accuracy. This surpasses not only the supervised ViT/B ($79.58\%$) but also the prior LLM-augmented supervised model, LM4Vision \citep{pang2023frozen} ($80.50\%$). More importantly, \ours outperforms the strong MAE-ViT/B baseline ($83.11\%$), demonstrating the impact of our synergistic LLM integration beyond standard MAE pretraining.
We note that \ours is much more stable between random seeds compared to LM4Vision \citet{pang2023frozen}, as quantified by the standard deviation between multiple runs.
We hypothesize that the gap between our reproductions and the values reported in \citet{pang2023frozen} could be attributed to these instabilities.

\myparagraph{\ours Better Leverages LLM Benefits.}
The MAE-pretrained ViT/B already provides a powerful visual backbone, outperforming the supervised ViT/B+LM1 ($83.11\%$ vs. $80.50\%$ on IN-1K). However, \ours builds on this strong foundation consistently and achieves respectable improvements. The improvements of \ours over the MAE-ViT baseline (e.g., $+0.52\%$ on IN-1K, $+2.75\%$ on IN-A) directly validate our hypothesis: concurrently training the LoRA-adapted LLM block during MAE pre-training enables the LLM to effectively process and enhance visual features. This joint optimization better bridges the modality mismatch, allowing the LLM to contribute semantic knowledge to the visual task, a benefit not realized by simply pre-training the ViT with MAE alone or even with extra  capacity as shown in Section~\ref{sec:ablation}.

\myparagraph{Enhanced Robustness and Generalization.}
The advantages of \ours become even more pronounced on robustness benchmarks. On IN-A, a particularly challenging adversarial dataset, \ours achieves a $\mathbf{13.24}\%$ improvement over LM4Vision \citep{pang2023frozen} and a $\mathbf{2.75}\%$ over the MAE-ViT baseline. \ours also attains respectable gains over both LM4Vision \citep{pang2023frozen} and MAE ViT baselines on IN-SK ($+3.13\%$, $+0.67\%$), IN-C ($+2.20\%$, $+0.58\%$), IN-V2 ($+4.46\%$, $+0.42\%$), and the visual attribute shifts of IN-E ($+0.54\%$). Crucially, as detailed in Appendix~\ref{app-sec:additional-results-imagenet-e}, \ours's advantage on IN-E significantly improves on the hardest splits, such as adversarial backgrounds ($+1.20\%$) and random spatial misalignments ($+1.24\%$), further solidifying the effectiveness of \ours.

With these results, the superior performance over the MAE-pretrained ViT shows that our method of integrating and adapting the LLM component brings tangible benefits beyond self-supervised visual pre-training. Second, it shows substantial improvements on robustness benchmarks (especially IN-A). The results indicate that \ours successfully leverages the LLM's knowledge to achieve improved resilience against out-of-distribution samples which is particularly important for real-world vision systems. Third, by outperforming previous attempts at LLM-ViT fusion, like LM4Vision~\citep{pang2023frozen}, \ours demonstrates the importance of both a strong pre-training paradigm (MAE) and an efficient adaptation strategy (concurrent LoRA training) to reap the benefits of the LLM block.

\subsection{Fine-grained Visual Recognition}
\label{sec:fine-grained-visual-recognition-exp}
We extend our evaluation beyond image classification to highlight \ours's capabilities in fine-grained visual recognition.
We conduct experiments on three diverse benchmarks: object detection and instance segmentation on MS COCO~\citep{lin2014mscoco}, video object segmentation on DAVIS-2017~\citep{pont20172017}, and semantic segmentation on Imagenet-Segmentation~\citep{gao2022large}.

\textbf{Object Detection and Instance Segmentation.}
We finetune \ours on MS COCO and compare it against the baselines. As shown in Table~\ref{tab:combined-fine-grained-results}, \ours consistently outperforms the strong MAE ViT baseline across all metrics.
Specifically, we achieve $51.1$ bounding box AP ($+0.5$ over MAE ViT/B). These improvements suggest that the co-adaptation strategy not only enhances global semantic understanding but also preserves the local spatial granularity required for precise localization.

\textbf{Video Object Segmentation.}
We further evaluate the temporal consistency and boundary precision of \ours's representations on DAVIS-2017. Table~\ref{tab:combined-fine-grained-results} summarizes these results.
\ours achieves a mean score $\mathcal{J}\&\mathcal{F}$ of $59.0$, significantly exceeding both the MAE ViT/B baseline ($57.3$) and the supervised-only LM4Vision ($57.2$).
A key observation is the substantial improvement in contour-based accuracy ($\mathcal{F}_M$), where \ours improves $+1.9$ over the MAE baseline.

\textbf{Unsupervised Object Segmentation.}
Finally, we evaluate the saliency of our features using the ImageNet-Segmentation-300 benchmark, reported in Table~\ref{tab:combined-fine-grained-results}.
Following the protocol of \citet{pang2023frozen}, we generate pseudo-masks from the frequency and magnitude components of the feature maps without any task-specific finetuning. \ours demonstrates superior zero-shot segmentation capabilities, achieving a frequency-based mIoU of $42.3$ ($+1.4$ over MAE ViT/B) and a magnitude-based mIoU of $43.5$.

Collectively, these results solidify that \ours successfully harnesses the LLM's semantic knowledge to improve visual recognition without compromising the fine-grained spatial precision essential for dense prediction.

\begin{table*}[t]
    \centering
    \caption{Performance comparison on MS COCO, DAVIS-2017 Video Object Segmentation, and ImageNet-Segmentation-300 (IN-Seg-300). For COCO, we report the Bounding Box AP. For DAVIS-2017, we report the mean Region Similarity ($\mathcal{J}_M$), Contour-based Accuracy ($\mathcal{F}_M$), and their average ($(\mathcal{J}\&\mathcal{F})_M$). For IN-Seg-300, we report mIoU. \textbf{Bold} denotes the best result.}
    \label{tab:combined-fine-grained-results}
    \scalebox{1.}{
    \begin{tabular}{c||c|c|c||c|c|c||c|c} 
    \toprule
    \midrule
    \multirow{2}{*}{Model} & \multicolumn{3}{c||}{COCO (Box)} & \multicolumn{3}{c||}{DAVIS-2017} & \multicolumn{2}{c}{IN-Seg-300} \\
     & $\text{AP}$ & $\text{AP}_{50}$ & $\text{AP}_{75}$ & $(\mathcal{J}\&\mathcal{F})_M$ & $\mathcal{J}_M$ & $\mathcal{F}_M$ & Freq. & Mag. \\
    \midrule
    
    MAE ViT/B & $50.6$ & $71.0$ & $55.5$ & $57.3$ & $56.4$ & $58.2$ & $40.9$ & $42.8$ \\
    \ours/B \textit{(Ours)} & $\mathbf{51.1}$ & $\mathbf{71.5}$ & $\mathbf{55.9}$ & $\mathbf{59.0}$ & $\mathbf{57.8}$ & $\mathbf{60.1}$ & $\mathbf{42.3}$ & $\mathbf{43.5}$ \\
     & $\imp{0.5}$ & $\imp{0.5}$ & $\imp{0.4}$ & $\imp{1.7}$ & $\imp{1.4}$ & $\imp{1.9}$ & $\imp{1.4}$ & $\imp{0.7}$ \\
    
    \midrule
    \bottomrule
    \end{tabular}}
\end{table*}

%% file: TMLR/Sections/4_3_Ablations.tex
\subsection{Ablations}
\label{sec:ablation}

\begin{table}[t]
\setlength{\tabcolsep}{0.25em}
\centering
\caption{Ablation analysis of \ours on ImageNet-1K show that \ours's design choices are essential to achieve the best performance. ``Train. Params." refers to parameters updated during fine-tuning, which includes the entire ViT, projections, and LoRA if present). ViT/B+MLP models are configured to match the trainable parameters of corresponding LLM-augmented models. Models are MAE pretrained and each result is the average of $3$ random seeds with their standard deviations. \textbf{Bold} indicates the best result.}
\label{tab:primary-ablations}

\scalebox{1.}{
\begin{tabular}{@{}l l|c||c@{}}
\toprule
\midrule
& Model & Train. Params. & IN-1K \\
\midrule
\textbf{(a)} & ViT/B & $86.8$M & $83.11_{\pm0.09}$ \\
\midrule
\textbf{(b)} & ViT/B+MLP-Proj. Match & $92.9$M & $83.13_{\pm0.06}$ \\
\textbf{(c)} & ViT/B+LM1 & $92.9$M & $83.13_{\pm0.02}$ \\
\midrule
\textbf{(d)} & ViT/B+MLP-LoRA Match & $93.1$M & ${83.21_{\pm0.11}}$ \\
\textbf{(e)} & ViT/B+Random LM1+LoRA & $93.1$M & ${83.25_{\pm0.09}}$ \\
\textbf{(f)} & \ours/B \textit{(Ours)} & $93.1$M & $\mathbf{83.63_{\pm0.04}}$ \\
\midrule
\bottomrule
\end{tabular}
}
\end{table}

\begin{table}[t]

    \centering
    \caption{Ablation analysis with different LLaMA 1 blocks and different LLMs' final blocks (LLaMA 1 7B, Gemma 2 9B, LLaMA 3.1 8B), show that \ours's improvements hold across different LLMs, with increasing improvements at the final blocks. All experiments had a random seed of $0$. \textbf{Bold} indicates the best result.}

    \label{tab:different-llm-ablation}
    \scalebox{1.}{
    \begin{tabular}{@{}c|l l|c||c@{}}
    \toprule
    \midrule
    && LLM Type & Block & IN-1K \\
    \midrule

    ViT/B && N/A & N/A & $83.2$ \\
    \midrule
    \multirow{7}{*}{\ours/B}
      &\textbf{(a)} & LLaMA 1 & $1$  & $83.2$ \\
      &\textbf{(b)} & LLaMA 1 & $16$ & $83.4$ \\
      &\textbf{(c)} & LLaMA 1 & $31$ & $83.5$ \\
      &\textbf{(d)} & LLaMA 1 (\textit{default}) & $32$ & $\mathbf{83.6}$ \\
    \cmidrule{2-5} 
      &\textbf{(e)} & Gemma 2 & $42$ & $83.5$ \\
      &\textbf{(f)} & LLaMA 3.1 & $32$ & $\mathbf{83.6}$ \\
      &\textbf{(g)} & LLaMA 3.1-Instruct & $32$ & $\mathbf{83.6}$ \\
    \midrule
    \bottomrule
    \end{tabular}
    }
\end{table}

In this section, we quantify the importance of the several building blocks of our approach: the pretrained LLM representations, the importance of LoRA and their combination with MAE pretraining.
We ablate these components and report the results on Tables~\ref{tab:primary-ablations} \& \ref{tab:different-llm-ablation} on ImageNet-1K.
We present more results in Section \ref{app-sec:additional-ablations-exp} on how \ours performs better compared to a baseline with a trainable LLM block and an ablation with $>1$ LLM blocks.

\myparagraph{LoRA Adaptation is Crucial for Leveraging LLM Benefits with MAE Pre-training.}
Comparing row (a) and (c) of Table~\ref{tab:primary-ablations}, we observe that the frozen LLM variant without any LoRA fine-tuning in row (c) ($83.13\%$ IN-1K) achieves on-par performance with the baseline MAE ViT of row (a) (row a: $83.11\%$ IN-1K). Without adaptation, the LLM block does not benefit from the richer features coming from the MAE-pre-training. This is in contrast with \citet{pang2023frozen} where the improvements were possible without LoRA on a weaker baseline.
However, when we introduce LoRA and adapt the LLM block, as in our full \ours/B model (row f), performance significantly improves to $\mathbf{83.63}\%$ on IN-1K. This is a clear improvement over both the MAE ViT/B baseline (row a) and the frozen LLM variant without LoRA (row c). Coupled with our study on multiple random seeds in Table \ref{tab:imagenet-main-results}, these results confirm that LoRA-based adaptation is \textit{essential} for effectively bridging the modality mismatch and allowing the LLM to use enhanced visual representations.

\myparagraph{\ours's Gains are Not Merely from Increased Parameters.}
A critical question is whether \ours's improvements stem from our model design or from an increased number of trainable parameters introduced by the linear projections and LoRA. To investigate this, we report additional results with two stronger baselines in Table \ref{tab:primary-ablations}, namely \textbf{(1) ViT/B+MLP (Proj. Match, row b)} and \textbf{(2) ViT/B+MLP (LoRA Match, row d)}.
The former's total trainable parameters ($92.9$M) match those of the ViT/B+LM1 (row c), which includes the ViT and the trainable linear projections, whereas the latter's total trainable parameters ($93.1$M) match those of our full \ours/B model (row f), which includes the ViT, trainable projections, and trainable LoRA layers.

Comparing row (b) with row (c), the ViT/B+MLP (Proj. Match) performs on-par on IN-1K compared to the frozen LLM without LoRA.
However, the crucial comparison is between our full \ours/B model (row f) and its parameter-matched MLP counterpart (row d). \ours/B achieves $\mathbf{83.63}\%$ on IN-1K, outperforming ViT/B+MLP (LoRA Match) (row d: $83.21\%$ IN-1K) by $+0.42\%$ on IN-1K.
Thus, these results quantify that the improvements of \ours are not simply due to additional training capacity but a direct consequence of our design choices.

\myparagraph{Pretrained LLM Features are Essential for \ours’s Gains.} Furthermore, to isolate the effects of architectural biases of appending an LLM block, we benchmarked another stronger baseline in Table \ref{tab:primary-ablations}, namely \textbf{ViT/B+Random LM1+LoRA (row e)}, identical to \ours architecturally, only with a randomly-initialized and LoRA-adapted LLaMA 1 block.
Echoing our observations from additional capacity ablations, ViT/B+Random LM1+LoRA only achieves on-par performance (row e: $83.25\%$ IN-1K) with ViT/B+MLP (LoRA Match), significantly falling behind \ours.

\myparagraph{\ours's Gains Are Robust with Different LLM Blocks and LLMs.}  We showcase how \ours's gains remain robustly high for a range of different LLM blocks in Table \ref{tab:different-llm-ablation}.
Particularly in Table \ref{tab:different-llm-ablation}, we replace the default LLM block (LLaMA 1, $32^{nd}$) of \ours with different blocks of LLaMA 1 and final blocks of different LLMs (Gemma 2 \citep{team2024gemma}, and LLaMA 3.1 \citep{grattafiori2024llama}).
Along with a clear trend of improvement as the block index gets closer to the final block (rows a-d), the performance of \ours remains largely invariant with the final blocks of different LLMs (rows e-g).
These results establish that the architectural biases alone cannot account for the performance gains of \ours, and the pretrained LLM representations are fundamental for the performance of \ours. 

%% file: TMLR/Sections/5_Analysis.tex
\section{On The Background Robustness of \ours}
\label{sec:analyses}
\begin{table*}[t]
    \centering
     \caption{Top-1 accuracy results of MAE pretrained models on Imagenet-9 adversarial backgrounds benchmark. The final three columns highlight the top-1 accuracy gap between different splits, a lower-better measure as denoted by the arrow $\downarrow$. \textbf{Bold} denotes best results.
     } 
    \label{tab:imagenet-9-results}
    \scalebox{1.}{
    \begin{tabular}{c||c|c|c|c|c|c} 
    \toprule
    \midrule
    Model&Original&Same&Random&\textit{Orig.-Same}$\downarrow$&\textit{Orig.-Rand.}$\downarrow$&\textit{Same-Rand.}$\downarrow$ \\ \midrule
    MAE ViT/B&${96.5}$&$87.8$&${83.2}$&$8.7$&$13.3$&${4.6}$ \\
    \ours/B \textit{(Ours)}&$\mathbf{96.6}$&$\mathbf{89.2}$&$\mathbf{85.3}$&$\mathbf{7.4}$&$\mathbf{11.3}$&$\mathbf{3.9}$\\
&$\imp{0.1}$&$\imp{1.4}$&$\imp{2.1}$&$\inimp{1.3}$&$\inimp{2.0}$&$\inimp{0.7}$\\
    \midrule
    \bottomrule
    \end{tabular}}
\end{table*}

\begin{figure}[t]
  \centering
  \scalebox{0.8}{
  \includegraphics[width=\linewidth]{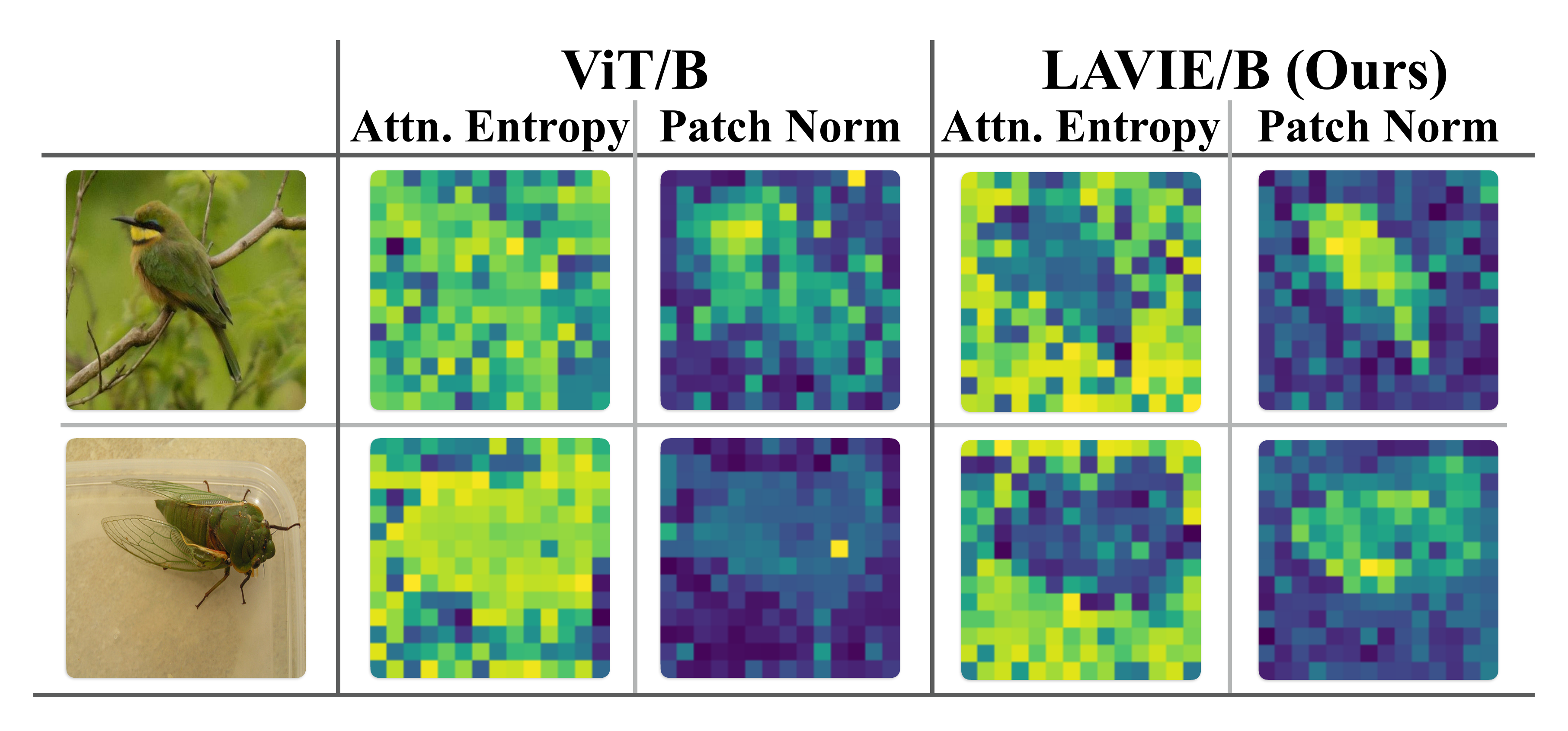}}
  \caption{
Visualized attention entropies and patch norms for \ours and the MAE-pretrained
ViT/B baseline. Across both examples, \ours exhibits lower attention entropy
and higher patch norms on foreground regions than ViT/B, indicating more
focused attention and improved saliency of foreground patch features. These
observations provide qualitative support for \ours's improved background
robustness. Brighter colors indicate higher values in the
\textbf{Attention Entropy} and \textbf{Patch Norm} columns.
}
  \label{fig:two-samples-visualization}
\end{figure}

\begin{figure}[t]
  \centering
  \begin{subfigure}[b]{0.4\linewidth}
    \centering
    \scalebox{1}{
    \includegraphics[width=\linewidth]{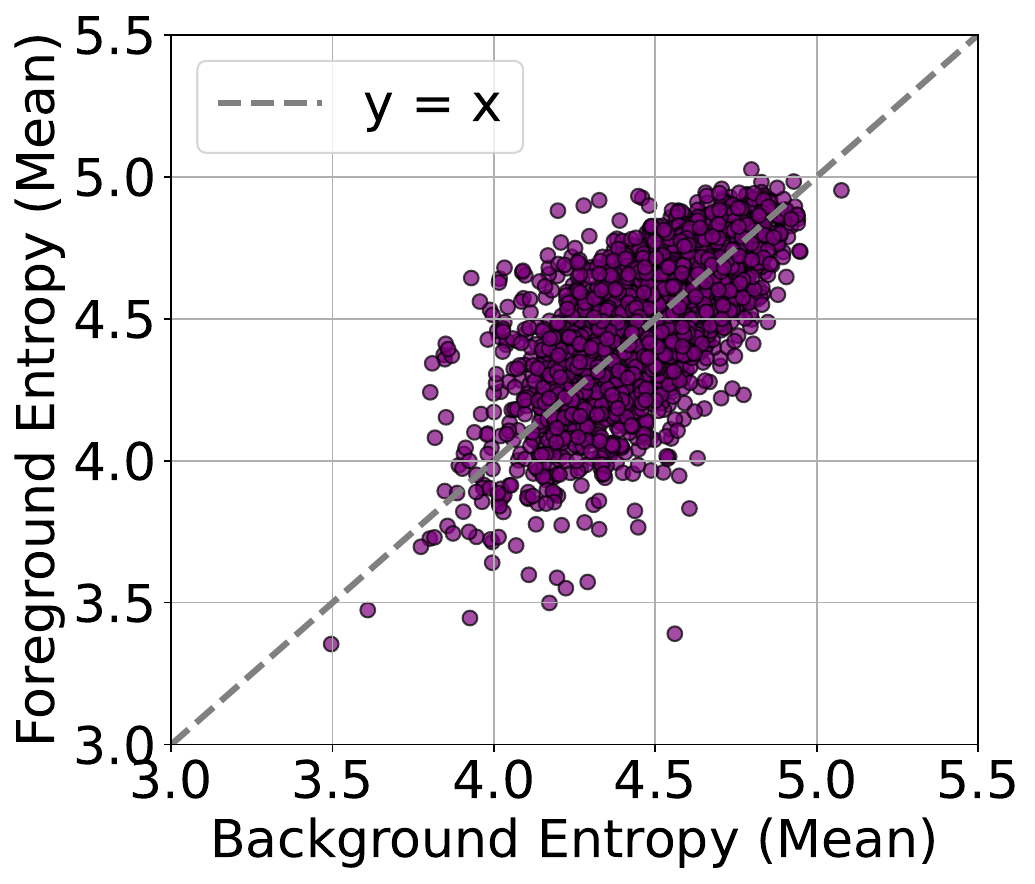}}
    \caption{ViT/B - Final Block Attention Entropies}
  \end{subfigure}
  \begin{subfigure}[b]{0.4\linewidth}
    \centering
    \scalebox{1}{
    \includegraphics[width=\linewidth]{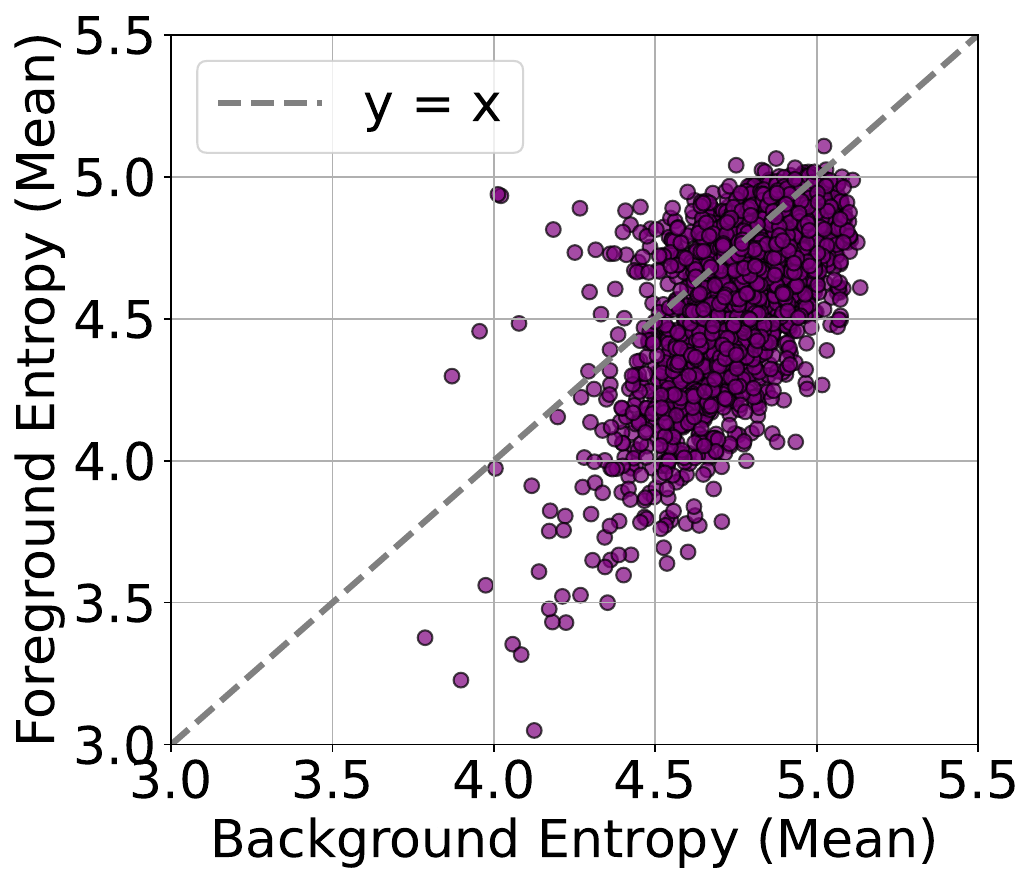}}
    \caption{\ours/B - Final Block Attention Entropies}
  \end{subfigure}
  \caption{Comparison of the image-level averaged foreground attention entropies vs background attention entropies of (a) MAE ViT/B baseline and (b) our \ours/B model. Each point in the plots corresponds to an image on Imagenet-S-300. \ours/B has a higher attention entropy for the backgrounds for $83\%$ of the images, and ViT/B has a higher attention entropy for the backgrounds for only $43\%$, resulting in ViT/B missing critical, information-rich foreground signals.}
  \label{fig:attention-entropy}
\end{figure}

In this section, we establish an intriguing connection between the background robustness and the improved performance by our \ours models, after analyzing the attention entropy patterns.
Previously, \citet{pang2023frozen} hypothesized that frozen LLM block could be acting as a filter, amplifying the final contributions of the informative tokens as part of their \textit{information filtering hypothesis}.
However, \citet{pang2023frozen} did not provide detailed discussions on the attention patterns, as they found the attention weights to be too noisy to provide insightful conclusions.
We aim to bridge this gap and providing deeper insights on \textit{how} \ours performs the \textit{information filtering}.

\myparagraph{\ours Exhibits More Focused Attention Patterns.}
We improve upon \citet{pang2023frozen}'s initial explorations and analyze the attention entropies of both the MAE ViT and our \ours, thereby decrypting the previously under-explored attention patterns of ViTs utilizing LLM blocks.
In particular, we quantify attention entropies through taking the post-softmax entropy of each row of the attention matrix, where each row corresponds to a spatial location on the feature map.%
 Formally, denoting the input as $X \in \mathbb{R}^{Txd}$, and the query and key projection matrices as $W_Q \in \mathbb{R}^{dxd_k}, W_K \in \mathbb{R}^{dxd_k}$, the post-softmax attention matrix with its row-wise entropies are given by:
\begin{equation}
\begin{gathered}
Q = XW_Q,\quad
K = XW_K,\quad
A = \operatorname{softmax}\!\left(\frac{QK^\top}{\sqrt{d_k}}\right), \\
H(A_i) = -\sum_{j=1}^{T} A_{i,j}\,\log A_{i,j}.
\end{gathered}
\label{eq:attn_entropy}
\end{equation}

We visualize the attention entropies on both image-level (Figure \ref{fig:two-samples-visualization}) and dataset-level on the Imagenet-S-300 dataset \citep{gao2022large} (Figure \ref{fig:attention-entropy}).
For the dataset-level visualizations, we map the mask annotations of Imagenet-S-300 down to the resolution of feature maps, and construct binary masks to distinguish the foreground regions from the background regions.
Then, we average the entropies of tokens belonging to the foreground vs background of each image.

In Figures \ref{fig:two-samples-visualization}\&\ref{fig:attention-entropy}, we observe a clear contrast between the attention entropies for the background and foreground regions for \ours, with the majority of the samples having significantly higher attention entropy for the background regions.
On the other hand, the average attention entropies are indifferent to background/foreground regions for the MAE ViT/B, highlighting its deficiency in differentiating informative regions from others.
Finally, we present more visualizations in the Appendix Section \ref{app-sec:visualizations}, where we further highlight the effective attention patterns of \ours.

\myparagraph{\ours is More Robust Against Adversarial Backgrounds.} Inspired by these observations in the attention patterns, we benchmark our \ours against the MAE ViT/B baseline on the challenging Imagenet-9, previously described in Section \ref{sec:experiments}.
Results in Table \ref{tab:imagenet-9-results} show that the gains of \ours significantly increase as the altered backgrounds become more challenging.
In particular, for \textit{Mixed Random} \& \textit{Mixed Same}, \ours improves the performance by $\mathbf{+2.1}$ and $\mathbf{+1.4}$.
Finally, \ours has significantly improved performance gaps between the original and background-altered splits, with gains reaching up to $\mathbf{2.0}$.

%% file: TMLR/Sections/6_Conclusion.tex
\section{Conclusion}

We introduce Language-Adapted Vision Enhancer (\ours), a framework that brings the semantic knowledge learned by text-only pre-trained LLM blocks into discriminative vision models. Our synergistic pre-training strategy leverages Masked Auto-Encoding (MAE) to learn rich visual representations from the ViT, while concurrently training LoRA layers within an LLM block using the same MAE objective. This joint optimization guides the ViT to produce LLM-friendly features and enables the LLM to enhance these visual features with its vast semantic knowledge.
Our comprehensive experiments demonstrate \ours's efficacy. In image classification benchmarks, \ours not only pushes the frontier under its setting but also shows greatly improved robustness to domain shifts compared to strong baselines.
Crucially, our analysis reveals this robustness stems from distinct attention dynamics: unlike standard ViTs, \ours learns to selectively minimize attention entropy on informative foreground regions, effectively filtering out background noise.
While MAE pre-training provides a strong foundation, the LoRA-based adaptation of the LLM block, trained in tandem, is essential for achieving performance gains. \ours offers a parameter-efficient pathway to harness the extensive knowledge of pre-trained LLMs for vision tasks.

%% file: TMLR/Supplementary_Material/A_0_Appendix.tex
\input{TMLR/Supplementary_Material/A_1_Visualizations}
\input{TMLR/Supplementary_Material/A_2_More_Experimental_Results.tex}
\input{TMLR/Supplementary_Material/A_3_Further_Details_on_Related_Work}
\input{TMLR/Supplementary_Material/A_4_Detailed_Training_Configs}

\input{TMLR/Supplementary_Material/A_5_Detailed_Datasets}
\input{TMLR/Supplementary_Material/A_6_Limitations}

%% file: TMLR/Supplementary_Material/A_1_Visualizations.tex
\section{More Visualizations with Attention Entropies}
\label{app-sec:visualizations}
\begin{figure*}[ht]
  \centering
  \scalebox{0.6}{
  \includegraphics[width=\linewidth]{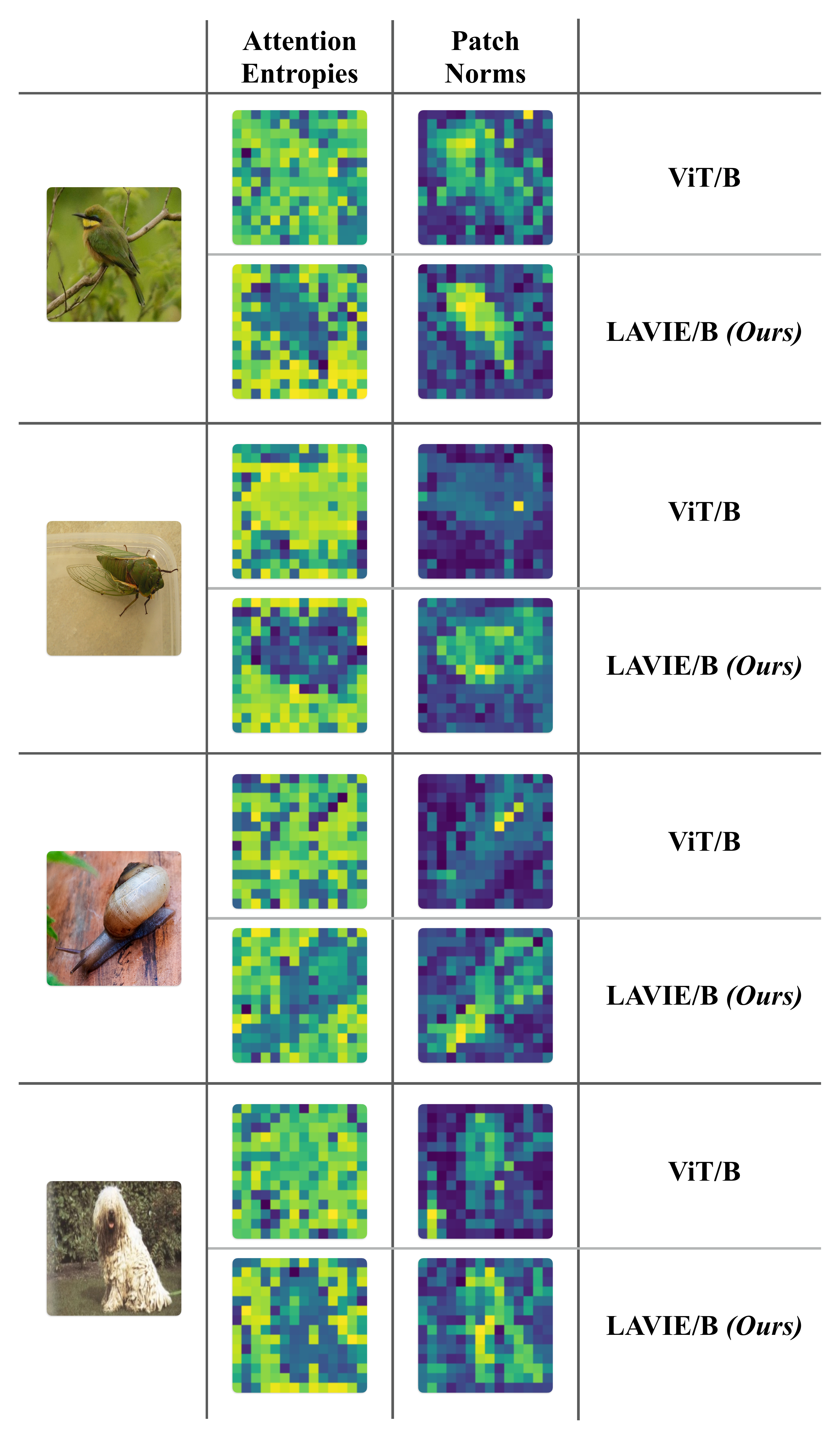}}
  \caption{Visualized attention entropies of both \ours/B and the MAE pre-trained ViT/B baseline.
  \ours/B simultaneously exhibits lower attention entropies and higher patch norms for foreground regions across all images compared to the ViT/B baseline, implying more focused attention patterns on these regions resulting in improved saliency in patch features.
  These results provide qualitative support to the background robustness behavior of \ours/B over the ViT/B baseline.
  The brighter colors highlight patches with high attention entropy, whereas the darker colors highlight patches with low attention entropy for the \textbf{``Attention Entropies''} column and the brighter colors highlight the patches with higher norm whereas the darker colors highlight the patches with lower norm for the \textbf{``Patch Norms''} column. }
  \label{fig:attention-entropy-images-local}
\end{figure*}
\begin{figure*}[t]
  \centering
  \scalebox{.95}{
  \includegraphics[width=\linewidth]{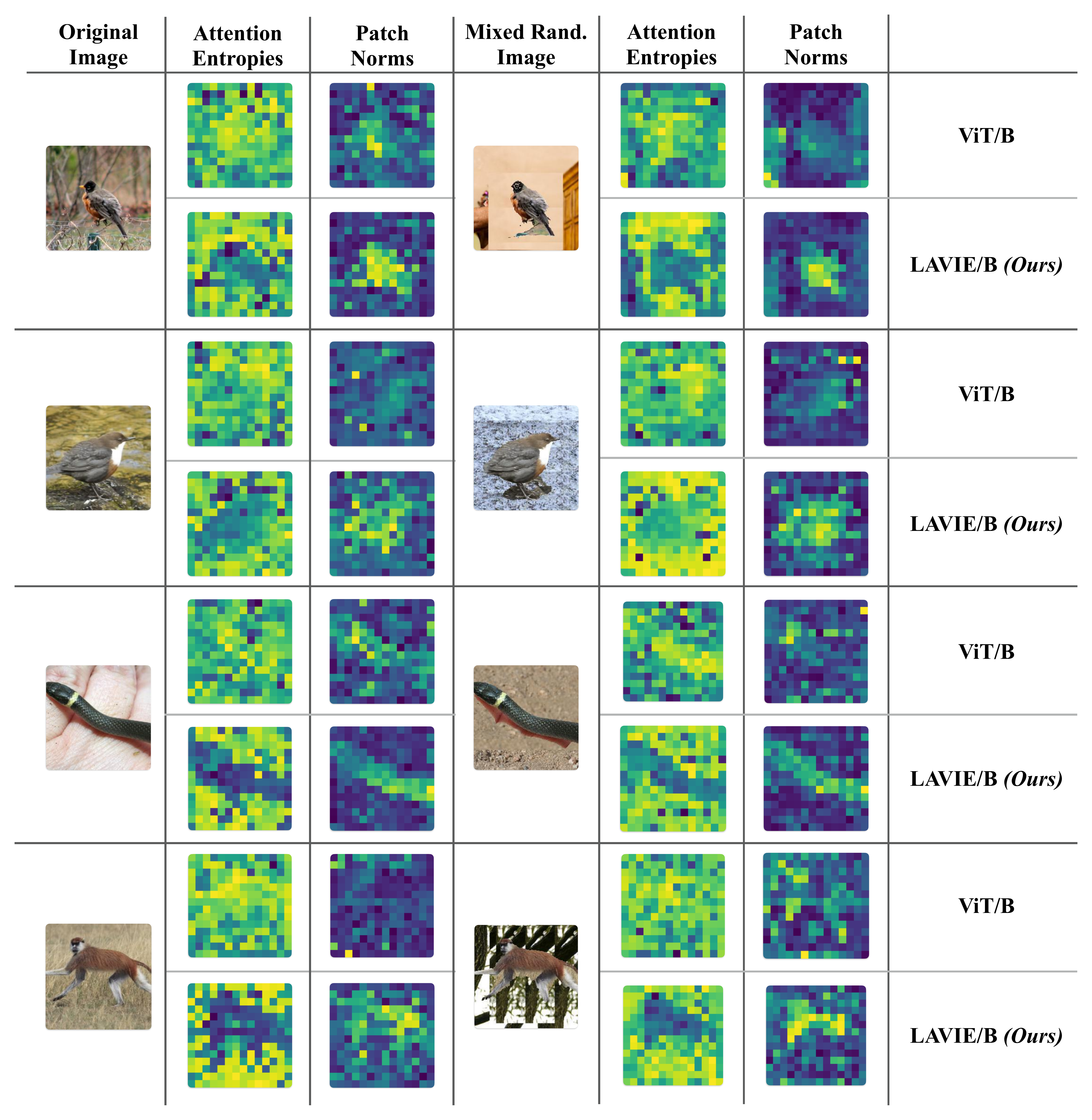}}
  \caption{Visualized attention entropies of both \ours/B and the MAE pre-trained ViT/B baseline on Imagenet-9 Dataset with \textit{Original} (denoted by column ``Original'') and \textit{Random} (denoted by column ``Mixed Rand.'') backgrounds.
  \ours/B simultaneously exhibits lower attention entropies and higher patch norms for foreground regions across all images compared to the ViT/B baseline, implying more focused attention patterns on these regions resulting in improved saliency in patch features.
  These results provide qualitative support to the background robustness behavior of \ours/B over the ViT/B baseline.
  The brighter colors highlight patches with high attention entropy, whereas the darker colors highlight patches with low attention entropy for the \textbf{``Attention Entropies''} column and the brighter colors highlight the patches with higher norm whereas the darker colors highlight the patches with lower norm for the \textbf{``Patch Norms''} column.}
  \label{fig:attention-entropy-images-local-in9}
\end{figure*}
In Sections 4 and 5, we experimentally demonstrated the effectiveness of \ours over the ViT baselines.
Furthermore, we provided in-depth analysis regarding the background robustness properties of \ours, where we demonstrated significant performance gains in Imagenet-9 \citep{xiao2020noise} with \ours under adversarial backgrounds .
Our empirical observations in Sections 4 and 5 were qualitatively grounded in the patterns we observe with the attention entropies of both \ours and the ViT baseline.
In particular, we showed that the foreground patches with \ours exhibit significantly lower attention entropy compared to background patches, whereas the same distinction does not occur with the baseline ViT.

With the aim of solidifying these observations, we provide additional visualizations of the attention entropy patterns for both our \ours and the baselines in this section.
The visualizations and results presented in this section demonstrate that both the attention entropy patterns and the patch norms for \ours provide significantly more salient visualizations compared to the ViT baseline (Section \ref{app-sec:image-level-visualizations}), and that the observations made from the scatter plots in Section \ref{sec:analyses} generalize across all splits of the Imagenet-Segmentation benchmark (Section \ref{app-sec:more-scatter-plots}).

\subsection{Image-level Attention Entropy Visualizations}
\label{app-sec:image-level-visualizations}
Here, we provide further details and image-level visualizations of attention entropy patterns along with the norms of the patches of both \ours and our MAE pre-trained ViT baselines in Figure \ref{fig:attention-entropy-images-local}.

Attention entropy patterns have been utilized in the context of neural network robustness in earlier works \citep{guo2023robustifying, zhang2024attention}.
In these works, they provided litmus tests for measuring how focused the attention patterns of particular models are and how they relate to model robustness.

As stated in Section \ref{sec:analyses}, we quantify the attention entropies through taking the post-softmax entropy of each row of the attention matrix, where each row corresponds to a spatial location, i.e., a patch, of the feature map, following the previous works using attention entropies \citep{zhai2023stabilizing}.

Formally, denoting the input as $X \in \mathbb{R}^{Txd}$, and the query and key projection matrices as $W_Q \in \mathbb{R}^{dxd_k}, W_K \in \mathbb{R}^{dxd_k}$, the post-softmax attention matrix with its row-wise entropies are given by:
\begin{equation}
\begin{gathered}
Q = XW_Q,\quad
K = XW_K,\quad
A = \operatorname{softmax}\!\left(\frac{QK^\top}{\sqrt{d_k}}\right), \\
H(A_i) = -\sum_{j=1}^{T} A_{i,j}\,\log A_{i,j}.
\end{gathered}
\label{eq:attn_entropy_app}
\end{equation}
Notably, we also average the attention entropies for each attention head, following the methodology of \citet{zhai2023stabilizing}.
Finally, to further supplement our visualizations, we additionally extract the L2 norms of each patch and visualize it alongside the attention entropy patterns.

Following this quantification process, we visualize the attention entropies along with the patch norms of both the final ViT block for both \ours and the MAE pre-trained ViT baseline after finetuning on Imagenet-1K \citep{deng2009imagenet} in Figures \ref{fig:attention-entropy-images-local} \& \ref{fig:attention-entropy-images-local-in9}.
For both Figures \ref{fig:attention-entropy-images-local} \& \ref{fig:attention-entropy-images-local-in9}, we perform a per-image normalization for both the patch norms and attention entropies to achieve more interpretable visualizations.
This corresponds to performing the normalizations based on the lowest and highest attention entropy score or token norm value for each feature map separately, and follows the normalization strategy used for visualizations in \citet{pang2023frozen}.

As it can be seen in Figure \ref{fig:attention-entropy-images-local}, \ours exhibits much lower attention entropies for the patches belonging to foreground regions compared to ViT/B, providing further qualitative support for our observations in Section \ref{sec:analyses}.
Simultaneously, the patch norms are more salient and achieve better coverage of foreground regions for \ours compared to ViT/B.
This behavior is specifically important, since we are utilizing average pooling instead of relying on the [CLS] token, following the default implementation in the official MAE codebase.
We refer the reader to Section \ref{app-sec:cls-e2e-details} for more details.

Furthermore, Figure \ref{fig:attention-entropy-images-local-in9} further demonstrates the behavioral changes under different Imagenet-9 settings for the same foreground objects.
Similar to Figure \ref{fig:attention-entropy-images-local}, across different settings, we observe that \ours exhibits much lower attention entropies for the patches belonging to foreground regions compared to ViT/B.
Here, we further note that the change of backgrounds also effects \ours and MAE ViT/B differently: While the behavior of \ours does not change drastically between the original image and the random background variant of the same image, the same does not hold for ViT/B, where we observe non-negligible artifacts in the form of high-norm patches in swapped background regions.

\subsection{Additional Attention Entropy Scatter Plots}
\label{app-sec:more-scatter-plots}
\begin{figure}[t]
  \centering
  \begin{subfigure}[b]{0.45\linewidth}
    \centering
    \scalebox{0.8}{
    \includegraphics[width=\linewidth]{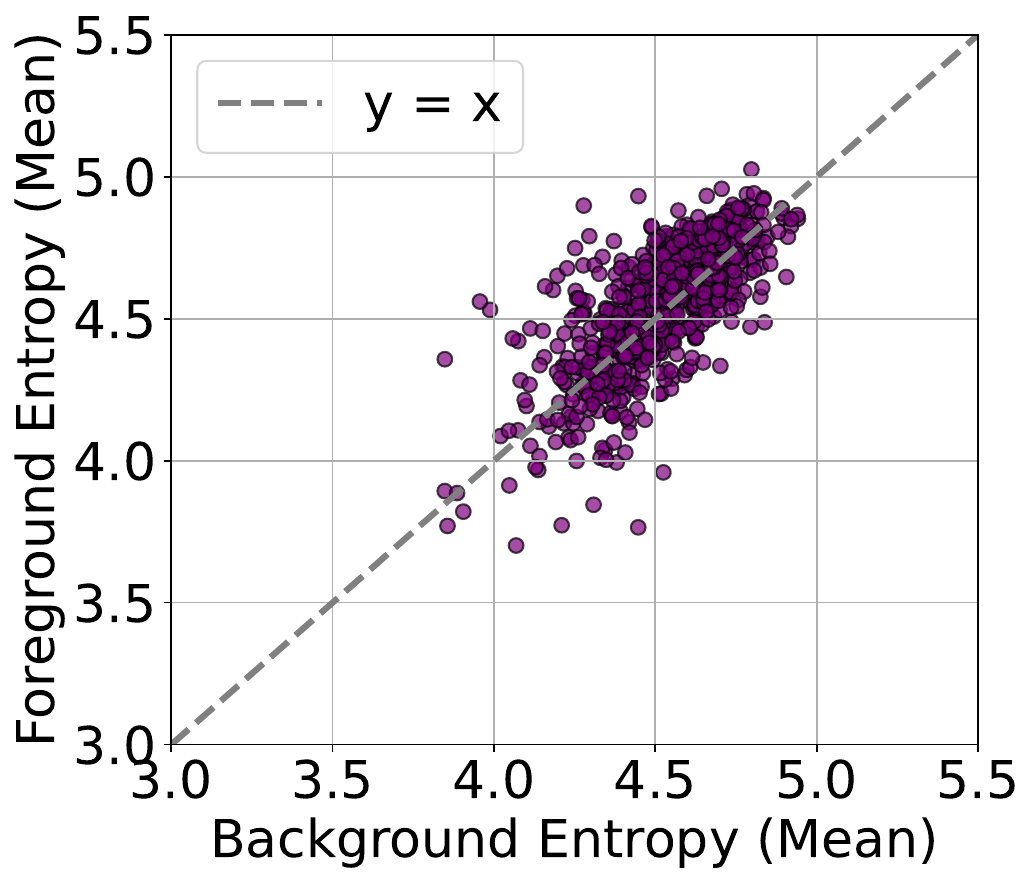}}
    \caption{ViT/B - Final Block Attention Entropies}
  \end{subfigure}
  \begin{subfigure}[b]{0.45\linewidth}
    \centering
    \scalebox{0.8}{
    \includegraphics[width=\linewidth]{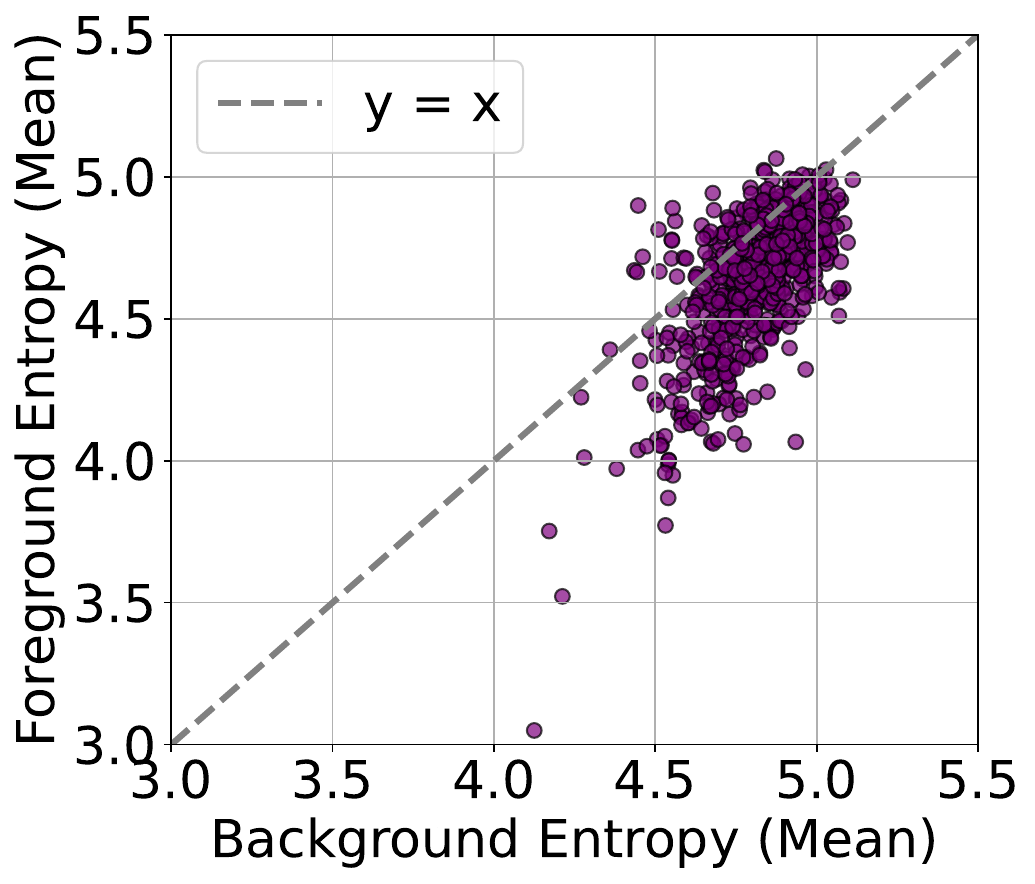}}
    \caption{\ours/B - Final Block Attention Entropies}
  \end{subfigure}

  \caption{Comparison of the image-level average foreground attention entropies vs the image-level average background attention entropies of (a) MAE ViT/B baseline and (b) our \ours/B model. Each point in the plots corresponds to an image on Imagenet-S-50 dataset \citep{gao2022large}. For $84\%$ of the images, \ours/B has a higher average attention entropy for the background regions, whereas this number drops to only $42\%$ for the ViT/B baseline.}
  \label{fig:attention-entropy-ins50}
\end{figure}

\begin{figure}[t]
  \centering
  \begin{subfigure}[b]{0.45\linewidth}
    \centering
    \scalebox{0.8}{
    \includegraphics[width=\linewidth]{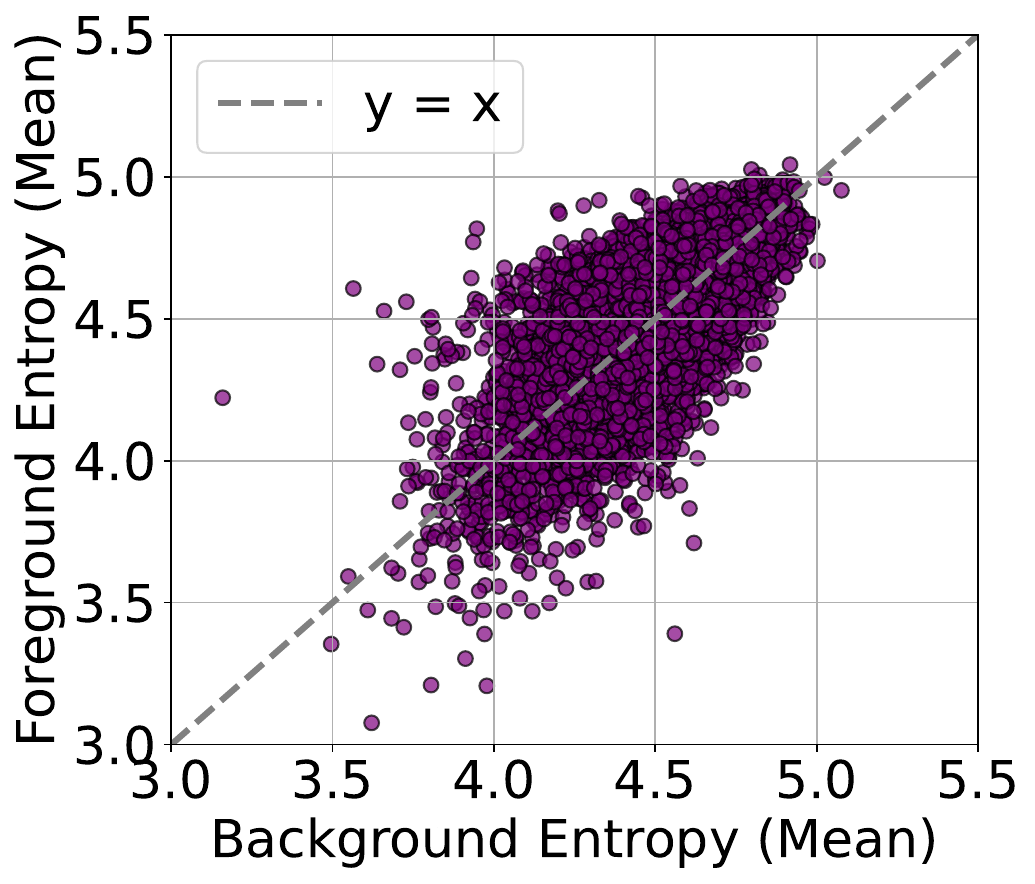}}
    \caption{ViT/B - Final Block Attention Entropies}
  \end{subfigure}
  \begin{subfigure}[b]{0.45\linewidth}
    \centering
    \scalebox{0.8}{
    \includegraphics[width=\linewidth]{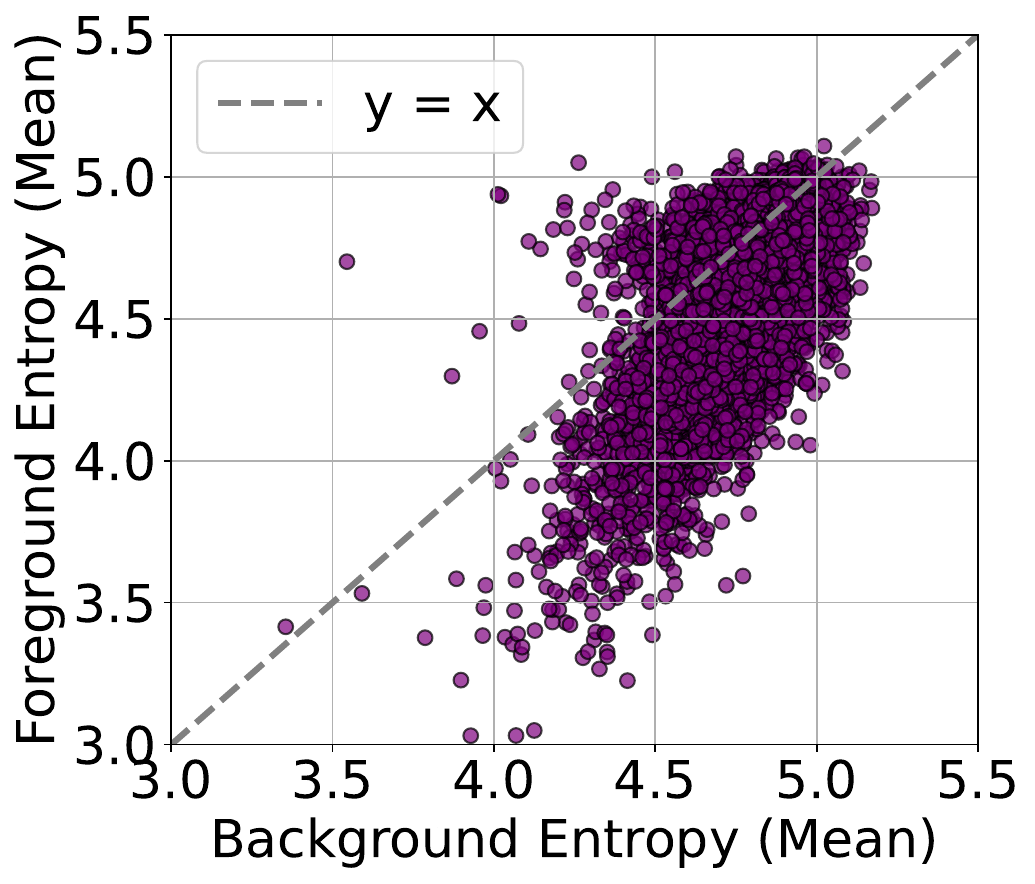}}
    \caption{\ours/B - Final Block Attention Entropies}
  \end{subfigure}

  \caption{Comparison of the image-level average foreground attention entropies vs the image-level average background attention entropies of (a) MAE ViT/B baseline and (b) our \ours/B model. Each point in the plots corresponds to an image on Imagenet-S-919 dataset \citep{gao2022large}. For $83\%$ of the images, \ours/B has a higher average attention entropy for the background regions, whereas this number drops to only $44\%$ for the ViT/B baseline.}
  \label{fig:attention-entropy-in919}
\end{figure}

In Section \ref{sec:analyses}, we presented the attention entropy scatter plots for the Imagenet-Segmentation-300 validation set \citep{gao2022large}.
Here, we additionally present the scatter plots for the other two Imagenet-Segmentation variants \citep{gao2022large}, namely for Imagenet-Segmentation-50 validation set in Figure \ref{fig:attention-entropy-ins50} and for Imagenet-Segmentation-919 validation set in Figure \ref{fig:attention-entropy-in919}.
Similar to the plots in Section \ref{sec:analyses}, each point in Figures \ref{fig:attention-entropy-ins50} and \ref{fig:attention-entropy-in919} correspond to the average attention entropy for each image where the y-axis highlights the average attention entropy for the foreground patches whereas the x-axis highlights the average attention entropy for the background patches.

These results closely mirror those in Section \ref{sec:analyses}, where again a very clear distinction emerges between the average attention entropies for the background and foreground regions for our \ours/B.
On the other hand, the attention entropies are mostly the same for all regions of the MAE pretrained ViT/B baseline, regardless of whether they belong to a highly informative foreground region or not.


%% file: TMLR/Supplementary_Material/A_2_More_Experimental_Results.tex
\section{Additional Experimental Results and Ablations}
In this section, we present
additional ablations several of our design choices (Section \ref{app-sec:additional-ablations-exp}), LoRA-adapting the LLM under supervised-only training (Section \ref{app-sec:supervised-only-llm-adaptation}), supplementary results on the Imagenet-Segmentation and DAVIS2017 benchmarks (Section \ref{app-sec:additional-results-imagenet-segmentation}) and detailed results on all Imagenet-E splits (Section \ref{app-sec:additional-results-imagenet-e}).

\subsection{Additional Ablations on Design Choices}
\label{app-sec:additional-ablations-exp}
In this section, we provide supplementary ablations over several different design choices.
Namely, Section \ref{app-sec:trainable-llm-block} shows that LoRA-adaptation can be more desirable to full finetuning of the LLM block under limited training set of Imagenet-1K, Section \ref{app-sec:two-llm-blocks} provides additional results using two LLaMA 1 blocks and finally Section \ref{app-sec:lora-rank-ablations} discusses the results under different ranks of LoRA.

\subsubsection{Adapting versus Full Finetuning the LLM Block}
\label{app-sec:trainable-llm-block}
\begin{table}[t]
    \centering
    \caption{Full finetuning of the LLM block can be undesirable under limited training sets, even though it has a much greater number of trainable parameters. ``ViT/B + Trainable LM1'' denotes the baseline with a fully-finetuned LLaMA 1 block. All results indicate Imagenet-1K top-1 accuracy. \textbf{Bold} denotes the better result.} 
    \vspace{1ex}
    \label{tab:trainable-llm-ablation}
    \begin{tabular}{c|c||c} 
    \toprule
    \midrule
    Model&Trainable Params.&Accuracy\\
    \midrule
    ViT/B &$86.8$M&$83.2$   \\
    ViT/B+Frozen LM1&$92.9$M&$83.1$ \\
    ViT/B+Trainable LM1&$295.2$M&$82.2$\\
    \ours (\textit{Ours})&$93.1$M&$\mathbf{83.6}$ ($\imp{0.4}$)\\

    \midrule
    \bottomrule
    \end{tabular}
\end{table}
Following the success of LoRA-adapting the LLM block in \ours, a natural question is what would the performance look like under full finetuning of the LLM block.
To address this question, we trained a baseline on Imagenet-1K where we left the LLM block completely trainable during both the MAE pretraining and end-to-end finetuning stages.
The results in Table \ref{tab:trainable-llm-ablation} closely mirror the observations of \citet{pang2023frozen}: The heavyweight LLM transformer block has a tendency to overfit, and regularizing it is very challenging, resulting in an even worse performance compared to the baseline with a completely frozen LLaMA 1 block.

\subsubsection{\ours with More Than One LLM Block}
\label{app-sec:two-llm-blocks}
\begin{table}[th]
    \centering
    \caption{Multiple LLM blocks could be promising, though they likely require a more careful hyperparameter selection. ``\ours 2-LM1'' denotes the baseline with two LoRA-adapted LLM blocks. All results indicate Imagenet-1K top-1 accuracy. \textbf{Bold} denotes the better result.} 
    \vspace{1ex}
    \label{tab:two-llm-blocks-ablation}
    \begin{tabular}{c|c||c} 
    \toprule
    \midrule
    Model&LLaMA 1 Blocks&Accuracy\\
    \midrule
    \ours 2-LM1&31 \& 32&$83.5$\\
    \ours (\textit{default})&32&$\mathbf{83.6}$\\

    \midrule
    \bottomrule
    \end{tabular}
\end{table}
Most of our explorations in Section \ref{sec:experiments} with \ours was around using only a single LLM block.
Another natural question is to ablate over the number of LLM blocks to be included to observe the possibility of reaping the benefits of the additional pretrained capacity.
To investigate this aspect, we trained another \ours variant with the final two LLaMA 1 blocks instead of only the final block.
We chose the final two blocks of LLaMA 1 following the results in Table \ref{tab:different-llm-ablation}, and report the results in Table \ref{tab:two-llm-blocks-ablation}.

The results in Table \ref{tab:two-llm-blocks-ablation} highlight that although having multiple LLM blocks is still significantly better than our previous baselines, it does not provide immediate performance improvements over our default configuration of using a single LLM block.
Following our quantitative observations of the two-blocks baseline simultaneously having higher training and testing errors, we hypothesize that its performance could be further improved with a more careful hyperparameter search, though we could not perform these experiments due to computational constraints.

\subsubsection{Ablations on LoRA Hyperparameters}
\label{app-sec:lora-rank-ablations}
\begin{table}[t]
    \centering
    \caption{Experiments with different LoRA hyperparameters highlight that rank and $\alpha$ $r=\alpha=16$ is sufficient enough for effectively adapting the LLM block. All results indicate Imagenet-1K top-1 accuracy. \textbf{Bold} denotes the best result.} 
    \vspace{1ex}
    \label{tab:lora-hyperparameters-ablation}
    \begin{tabular}{c|c||c} 
    \toprule
    \midrule
    Model&LoRA Rank&Accuracy\\
    \midrule
    \multirow{3}{*}{\ours}& $r=\alpha=8$&$83.3$\\
    &$r=\alpha=16$ (\textit{default})&$\mathbf{83.6}$\\
    &$r=\alpha=32$&$\mathbf{83.6}$\\

    \midrule
    \bottomrule
    \end{tabular}
\end{table}
In this section, we present additional results with different rank and $\alpha$ hyperparameters for the LoRA layers of \ours.
The results in Table \ref{tab:lora-hyperparameters-ablation} show that setting $r=\alpha=16$ is sufficient for achieving the performance gains, since the performance seems to saturate beyond this rank, as evidenced by the similar performances of the $r=\alpha=16$ and  $r=\alpha=32$ configurations.
Furthermore, the $r=\alpha=8$ seems to be insufficient for fully reaping the benefits of the pretrained LLM representations.
Accordingly, we opted for $r=\alpha=16$ for our experiments in Section \ref{sec:experiments}.

\subsubsection{Ablations on RoPE and Causal Attention Inside the LLM Block}
\label{app-sec:llm-rope-causal-ablations}
\begin{table}[t]
    \centering
    \caption{Experiments with keeping and turning off RoPE and causal attention inside the LLM block highlight that turning off these two components is crucial for achieving the performance gains. All results indicate Imagenet-1K top-1 accuracy. \textbf{Bold} denotes the best result.} 
    \vspace{1ex}
    \label{tab:llm-rope-causal-ablation}
    \begin{tabular}{c|l||c} 
    \toprule
    \midrule
    Model&LoRA Rank&Accuracy\\
    \midrule
    \multirow{2}{*}{\ours}& with RoPE \& Causal Attn.&$83.3$\\
    &without RoPE \& Causal Attn. (\textit{default})&$\mathbf{83.6}$\\
    \midrule
    \bottomrule
    \end{tabular}
\end{table}
As introduced in Section 3.3, a core architectural adjustment in \ours is the removal of the 1D Rotary Positional Embeddings (RoPE) and the replacement of the causal attention mask with bidirectional attention.
To empirically validate these design choices, we conduct an ablation study where we retain both the original 1D RoPE and the causal attention masking inside the frozen LLaMA 1 block.
As shown in Table \ref{tab:llm-rope-causal-ablation}, preserving these text-centric structural biases results in a performance degradation, dropping the ImageNet-1K top-1 accuracy from $83.6\%$ (our default configuration) to $83.3\%$.
This drop demonstrates that without these structural modifications, the pre-trained LLM representations are rendered less effective for visual inputs.
Unlike language, visual information within an image does not possess a strict, sequential causal structure.
Consequently, causal masking unnecessarily restricts visual tokens from attending to the full global context.
Furthermore, 1D RoPE enforces sequential relative positional biases tailored for text sequences, which disrupts the spatial relationships of image patches already captured by the ViT's learned positional embeddings.
Therefore, by converting to bidirectional attention and removing RoPE, we successfully overcome language-specific constraints, ensuring ensures that the LLM block can effectively map its high-level semantic priors to the spatially unconstrained visual domain.

\subsubsection{Ablations on the Placement of the LLM Block Within the ViT}
\label{app-sec:vit-block-insertion-ablations}
\begin{table}[t]
\centering
\caption{Ablation study on the placement of the LLM block within the ViT. Placing the LLM block deeper in the network yields better performance, with the highest accuracy achieved when placed after the final block. All results indicate ImageNet-1K Top-1 accuracy (\%). \textbf{Bold} denotes the best result.}
\vspace{1ex}
\label{tab:llm-placement-ablation}
\begin{tabular}{c|l||c} 
\toprule
\midrule
Model & LLM Placement & Accuracy \\
\midrule
\multirow{3}{*}{\ours} & After 1st block & $83.3$ \\
& After 6th block & $83.4$ \\
& After 12th (final) block (\textit{default}) & $\mathbf{83.6}$ \\
\midrule
\bottomrule
\end{tabular}
\end{table}

In our default \ours architecture introduced in Section \ref{sec:methodology}, the LLM fusion block is integrated immediately after the final block of the ViT.
To validate this structural design choice, we conduct an ablation study investigating the effect of placing the LLM block at different depths within the ViT backbone.
Specifically, we evaluate the end-to-end performance of \ours when the LLM block is inserted after the 1st, 6th, and 12th (final) blocks of the ViT encoder.

The results, presented in Table \ref{tab:llm-placement-ablation}, reveal a clear trend: placing the LLM block deeper within the network consistently yields better performance, culminating in the highest top-1 accuracy of $83.6\%$ at the 12th block.
This behavior strongly aligns with established findings regarding the feature hierarchy in Vision Transformers \citep{dosovitskiy2020image, raghu2021vision}.
Early layers of a ViT typically capture localized, low-level visual features (e.g., edges and simple textures), whereas deeper layers construct highly global, semantically separable representations. 
Accordingly, we hypothesize that because the pre-trained LLM block is optimized to operate within a highly abstract, semantic latent space, it struggles to effectively process raw, low-level visual tokens, resulting in a sub-optimal accuracy of $83.3\%$ when placed after the 1st block.
Conversely, inserting the LLM block at the end of the ViT ensures that the visual features have reached a sufficient level of semantic maturity.
This allows the LoRA-adapted LLM block to effectively interpret the visual tokens and seamlessly fuse them with its extensive language priors, maximizing the downstream discriminative performance.
\subsection{Adapting the LLM Under Supervised-only Training}
\label{app-sec:supervised-only-llm-adaptation}
In Section \ref{sec:experiments}, we demonstrated the effectiveness of employing Masked Auto-Encoding (MAE) to pre-
train the ViT for richer visual representations, while concurrently training Low-Rank Adaptation (LoRA) layers within the LLM block using the same MAE objective.

Following our results and ablations in Section \ref{sec:analyses}, a natural question may arise regarding how well a concurrent training strategy of Low-Rank Adaptation (LoRA) layers, and the ViT could work \textit{without} the critical MAE pretraining phase.
In this section, we investigate this question by including Low-Rank Adaptation (LoRA) layers on top of the architecture proposed in \citet{pang2023frozen} in a supervised-only setting.

The results of both our reproduction of \citet{pang2023frozen}'s model and the LoRA-adapted version of it are presented in Table \ref{tab:supervised-only-lora-ablation}.
For Table \ref{tab:supervised-only-lora-ablation}, we train both \citet{pang2023frozen}'s LM1+ViT/B and its LoRA-adapted version in a supervised-only setting on Imagenet-1K with three random seeds 
while adhering to all training settings in \citet{pang2023frozen}.
Furthermore, we directly utilize their code-base \footnote{\url{https://github.com/ziqipang/LM4VisualEncoding}}, and merely inject trainable LoRA layers to its LLM block, similar to the methodology of \ours presented in Sections 3 and 4.
Finally, we report the average accuracy across the seeds with the accompanying standard error values, i.e. the standard deviation of the accuracy values divided by the number of different seeds.

From Table \ref{tab:supervised-only-lora-ablation} we observe that the LoRA version achieves a slightly improved performance compared to the LM1+ViT/B. Concretely, the LoRA-adapted version of Pang et al. [2023]’s LM1+ViT/B has a
+0.12 better accuracy compared to the completely frozen LM1+ViT/B.
These results highlight that LoRA could potentially address the training instabilities that arise when directly finetuning the LLM block even under a supervised-only setting, though the performance improvements are much less pronounced compared to \ours’s improvements over the MAE ViT/B baselines.

\begin{table}[t]
    \centering
    \caption{Adapting the LLM block in the ViT of \citet{pang2023frozen} with a supervised-only training regime. Each reported value is an average runs with three random seeds and the subscript $\pm$ denotes the standard error for each setting. Note that we report two significant digits in the decimal for highlighting the effect of standard errors in contrast with other tables. \textbf{Bold} denotes the best result.} 
    \vspace{1ex}
    \label{tab:supervised-only-lora-ablation}
    \begin{tabular}{c||c} 
    \toprule
    \midrule
    Model&Average Accuracy\\
    \midrule
    LM1+ViT/B&$80.51_{\pm 0.07}$ \\
    LoRA LM1+ViT/B&$80.63_{\pm 0.09}$\\
    \midrule
    \bottomrule
    \end{tabular}
\end{table}

\subsection{Additional Results on Imagenet-Segmentation and DAVIS2017 Benchmarks}
\label{app-sec:additional-results-imagenet-segmentation}
In this section, we provide supplementary quantitative evidence expanding upon the dense prediction experiments from Section \ref{sec:fine-grained-visual-recognition-exp} and the qualitative token norm observations from Figure \ref{fig:attention-entropy-images-local}. We achieve this by presenting results across multiple random seeds for the DAVIS-2017 \citep{pont20172017} and ImageNet-Segmentation \citep{gao2022large} benchmarks.

\paragraph{Video Object Segmentation on DAVIS-2017.}
Table \ref{tab:davis-results} presents the evaluation of \ours on the DAVIS-2017 benchmark across three random seeds. This task heavily relies on precise boundary delineation and temporal consistency. \ours consistently outperforms the strong MAE ViT/B baseline across all metrics. Most notably, \ours achieves a mean Region Similarity ($\mathcal{J}_M$) and Contour-based Accuracy ($\mathcal{F}_M$) average of $60.17\%$, marking a robust $+2.03\%$ improvement over the baseline. The non-overlapping standard deviations strongly indicate that the enhanced spatial granularity driven by the LoRA-adapted LLM block is both statistically significant and highly stable.

\begin{table}[t]
    \centering
    \caption{Performance comparison on DAVIS-2017 Video Object Segmentation. We report the mean Region Similarity ($\mathcal{J}_M$), Contour-based Accuracy ($\mathcal{F}_M$), and their average ($(\mathcal{J}\&\mathcal{F})_M$). Each result denotes the average of $3$ random seeds along with associated standard deviations. \textbf{Bold} denotes the best result.}
    \label{tab:davis-results}
    \scalebox{1.}{
    \begin{tabular}{c||c|c|c} 
    \toprule
    \midrule
    \multirow{2}{*}{Model} & \multicolumn{3}{c}{DAVIS-2017} \\
     & $(\mathcal{J}\&\mathcal{F})_M$ & $\mathcal{J}_M$ & $\mathcal{F}_M$ \\
    \midrule
    MAE ViT/B & $58.13_{\pm0.80}$ & $57.17_{\pm0.80}$ & $59.03_{\pm0.76}$ \\
    \ours/B \textit{(Ours)} & $\mathbf{60.17_{\pm1.01}}$ & $\mathbf{59.20_{\pm1.22}}$ & $\mathbf{61.10_{\pm0.87}}$ \\
     & $\imp{2.03}$ & $\imp{2.03}$ & $\imp{2.07}$ \\
    \midrule
    \bottomrule
    \end{tabular}}
\end{table}

\paragraph{Unsupervised Saliency on ImageNet-Segmentation.}
Next, we compare the binary mask Intersection over Union (IoU) of the patch features with respect to the ground truth segmentation masks across all three ImageNet-Segmentation splits in Table \ref{tab:ious}. Following the methodology of \citet{pang2023frozen}, we extract the \textbf{magnitude} and \textbf{frequency} components from the token features to generate zero-shot pseudo-masks without task-specific finetuning.
Specifically, the magnitude component is obtained by taking the L2 norm of each centered token feature vector, while the frequency component is derived via a Fast Fourier Transform (FFT) \citep{pang2023frozen}. Binary pseudo-masks are then created by applying an empirically determined fixed threshold to these values. The ground-truth segmentation masks are downsampled to match the $14^2$ resolution of the ViT/B feature maps. For further details on the frequency and magnitude extractions, we refer the reader to Appendix A.3 of \citet{pang2023frozen}.

As evidenced in Table \ref{tab:ious}, \ours demonstrates higher IoUs across all three data subsets and across both the frequency and magnitude components. Once again, evaluating over three random seeds confirms the stability of these gains. Notably, \ours achieves an average IoU gain of $+1.50\%$ for the frequency component on the challenging INS-919 split. These improvements directly support our qualitative findings, confirming that the \ours architecture inherently forms more salient, background-resistant patch features than the standard MAE pretrained ViT baseline.

\begin{table}[t]
\setlength{\tabcolsep}{0.4em}
    \centering
    \caption{Mask IoUs of the final ViT block for each model with respect to the Imagenet-Segmentation \citep{gao2022large} annotations across different splits (INS-50, INS-300, INS-919). The Frequency column shows the results when the frequency component of the token features is used for obtaining the binary masks, whereas the Magnitude column shows the results when the magnitude component is used. Each result denotes the average of $3$ random seeds along with associated standard deviations. \textbf{Bold} denotes the best result.} 
    \vspace{1ex}
    \label{tab:ious}
    \begin{tabular}{c||c|c|c|c|c|c} 
    \toprule
    \midrule
    \multirow{2}{*}{Model}&\multicolumn{2}{c|}{INS-50}&\multicolumn{2}{c|}{INS-300}&\multicolumn{2}{c}{INS-919}\\
    &Frequency&Magnitude&Frequency&Magnitude&Frequency&Magnitude\\
    \midrule
    MAE ViT/B & $39.73_{\pm0.12}$ & $42.37_{\pm0.06}$ & $40.87_{\pm0.06}$ & $42.80_{\pm0.10}$ & $40.37_{\pm0.67}$ & $42.63_{\pm0.21}$ \\
    \ours/B \textit{(Ours)} & $\mathbf{40.97_{\pm0.29}}$ & $\mathbf{42.77_{\pm0.09}}$ & $\mathbf{42.03_{\pm0.23}}$ & $\mathbf{43.23_{\pm0.23}}$ & $\mathbf{41.87_{\pm0.21}}$ & $\mathbf{43.03_{\pm0.23}}$ \\
     & $\imp{1.23}$ & $\imp{0.40}$ & $\imp{1.17}$ & $\imp{0.43}$ & $\imp{1.50}$ & $\imp{0.40}$ \\
    \midrule
    \bottomrule
    \end{tabular}
\end{table}

\subsection{Detailed Results on Imagenet-E Benchmark}
\label{app-sec:additional-results-imagenet-e}
\begin{table*}[t]
\centering
\caption{\ours achieves significantly better Top-1 accuracy (\%) on the \textbf{background} splits of the ImageNet-E dataset in a frozen LLM augmented model setting. We evaluate against variants of background texture complexity ($\lambda=-20$, $\lambda=20$), adversarial backgrounds (Adv), and random backgrounds (Rand). \ours consistently outperforms both supervised baselines and the strong MAE-pretrained ViT/B. Each result denotes the average of $3$ random seeds along with associated standard deviations. \textbf{Bold} indicates the best result.}

\label{tab:imagenet-e-bg-results}

\scalebox{.95}{
\begin{tabular}{c|c||c|c|c|c|c}
\toprule
\midrule
Training & Model & Original & -20 & 20 & Adv & Rand \\
\midrule
Supervised-Only & ViT/B &
$92.50_{\pm1.20}$ & $86.67_{\pm1.70}$ & $83.47_{\pm1.82}$ & $63.83_{\pm3.16}$ & $79.53_{\pm2.00}$ \\
{[LM4Vision]} & ViT/B+LM1 &
$93.03_{\pm0.23}$ & $86.83_{\pm0.51}$ & $84.00_{\pm0.20}$ & $63.33_{\pm0.55}$ & $80.20_{\pm0.53}$ \\
\citep{pang2023frozen} & & & & & & \\
\midrule
MAE Pretrained & ViT/B &
$95.23_{\pm0.06}$ & $90.70_{\pm0.20}$ & $89.57_{\pm0.35}$ & $73.90_{\pm0.20}$ & $87.43_{\pm0.35}$ \\
& \ours/B \textit{(Ours)} &
$\mathbf{95.30_{\pm0.10}}$ & $\mathbf{91.13_{\pm0.06}}$ & $\mathbf{89.77_{\pm0.21}}$ & $\mathbf{75.10_{\pm0.20}}$ & $\mathbf{87.87_{\pm0.35}}$ \\
& &
$\imp{0.07}$ & $\imp{0.43}$ & $\imp{0.20}$ & $\imp{1.20}$ & $\imp{0.44}$ \\
\midrule
\bottomrule
\end{tabular}
}
\end{table*}
\begin{table*}[t]
\centering
\caption{\ours achieves significantly better Top-1 accuracy (\%) on the \textbf{size, position, and direction} splits of the ImageNet-E dataset in a frozen LLM augmented model setting. We evaluate against variants of object size (0.1, 0.08, 0.05 pixel rates), random position (rp), and random direction (rd). \ours consistently outperforms both supervised baselines and the strong MAE-pretrained ViT/B. Each result denotes the average of $3$ random seeds along with associated standard deviations. \textbf{Bold} indicates the best result.}
\label{tab:imagenet-e-other-results}

\scalebox{.87}{
\begin{tabular}{c|c||c|c|c|c|c|c}
\toprule
\midrule
Training & Model & Original & 0.1 & 0.08 & 0.05 & rp & rd \\
\midrule
Supervised-Only & ViT/B &
$92.50_{\pm1.20}$ & $88.13_{\pm1.96}$ & $85.23_{\pm2.40}$ & $75.73_{\pm3.25}$ & $70.67_{\pm3.06}$ & $76.00_{\pm1.37}$ \\
{[LM4Vision]} & ViT/B+LM1 &
$93.03_{\pm0.23}$ & $88.77_{\pm0.15}$ & $85.83_{\pm0.40}$ & $76.63_{\pm0.51}$ & $72.40_{\pm0.60}$ & $76.93_{\pm0.67}$ \\
\citep{pang2023frozen} & & & & & & & \\
\midrule
MAE Pretrained & ViT/B &
$95.23_{\pm0.06}$ & $92.60_{\pm0.26}$ & $90.73_{\pm0.31}$ & $83.53_{\pm0.06}$ & $79.23_{\pm0.21}$ & $81.40_{\pm0.35}$ \\
& \ours/B \textit{(Ours)} &
$\mathbf{95.30_{\pm0.10}}$ & $\mathbf{93.17_{\pm0.38}}$ & $\mathbf{91.30_{\pm0.17}}$ & $\mathbf{84.30_{\pm0.26}}$ & $\mathbf{80.47_{\pm0.40}}$ & $\mathbf{81.43_{\pm0.31}}$ \\
& &
$\imp{0.07}$ & $\imp{0.57}$ & $\imp{0.57}$ & $\imp{0.77}$ & $\imp{1.24}$ & $\imp{0.03}$ \\
\midrule
\bottomrule
\end{tabular}
}
\end{table*}

Table \ref{tab:imagenet-e-bg-results} details the performance across different background perturbations, including variations in texture complexity ($\lambda=-20, 20$), adversarial backgrounds (Adv), and random backgrounds (Rand). \ours consistently outperforms the MAE ViT/B baseline, with the most substantial gains observed in the highly challenging adversarial background split ($+1.20\%$).

Table \ref{tab:imagenet-e-other-results} presents the results for geometric and spatial attribute shifts, specifically varying object sizes ($0.1$, $0.08$, $0.05$ pixel rates), random positions (rp), and random directions (rd). Similarly, \ours demonstrates superior robustness, particularly when the object is shifted to a random position ($+1.24\%$) or extremely downscaled to a $0.05$ pixel rate ($+0.77\%$).

Collectively, these fine-grained ImageNet-E results reinforce the core hypothesis of our work: the semantic priors of the pretrained LLM block do not just marginally improve clean accuracy, but act as a powerful regularizer that anchors the model's predictions when the visual stream is heavily degraded or spatially distorted.

%% file: TMLR/Supplementary_Material/A_3_Further_Details_on_Related_Work.tex
\section{Further Discussion of Related Work}
\label{app-sec:further-related-work}
In this section, we discuss the works most closely related to ours in greater detail while highlighting the key differences, similarities, and orthogonal directions between them and our \ours framework.

\subsection{Information Filtering Hypothesis}
An important contribution of \citet{pang2023frozen} was the introduction of the \textit{information filtering hypothesis}.
The information filtering hypothesis was proposed as a potential explanation for how a frozen LLM block could enhance visual features for visual recognition tasks.
In particular, \citet{pang2023frozen} build on the DeiT \citep{touvron2021training} family of models and perform classification using the [CLS] token.
The authors then argued that, to achieve better performance than the vanilla ViT/B, either the attention weights should improve or the informative tokens should be amplified by the LLM block.

Formally, let $V$ denote the set of visual tokens and $w_v$ the attention weight of visual token $v$ in the final ViT block.
We denote the visual token processed by the first linear layer following the ViT block as $M^1_L(z[v]) = z^1_v[v]$, and the [CLS] token processed by the LLM block as $z'_{\text{[CLS]}}$.
The hypothesis proposes the following correlation:
\begin{equation}
    z'_{\text{[CLS]}} \propto \sum_{v \in V} w_v (M_L^2 \cdot M_{LLM} \cdot z_v^1[v]),
\end{equation}
under the assumption that $M_L^2 \cdot M_{LLM} \cdot M_L^1$ is a linear projection.

However, \citet{pang2023frozen} made the qualitative observation that the attention weights, $w_v$, were noisy, thus concluding that the $M_L^2 \cdot M_{LLM}$ projection must amplify the most informative tokens.

While \ours differs from the frozen-LLM-appended ViTs of \citet{pang2023frozen} in several key architectural and training-related details,
our work also leverages pre-trained LLM representations to improve discriminative visual recognition tasks.
In addition, as discussed in Section~\ref{sec:analyses} and Section~\ref{app-sec:visualizations}, \ours exhibits strong robustness against adversarial backgrounds compared with the baselines, a potential consequence of the information filtering hypothesis.
Coupled with our attention-entropy observations, the analyses presented in Sections~\ref{sec:analyses} and~\ref{app-sec:visualizations} can be understood in the same spirit as the information filtering hypothesis and provide complementary evidence.
\subsection{Pretrained LLM Layers and Gradient Coherence in Vision Transformers}
Another recent work investigating the mechanisms through which a frozen LLM block improves visual recognition performance is that of \citet{bai2025frozen}, who approach the question from a gradient-dynamics perspective.

In particular, \citet{bai2025frozen} broadly adopted the architecture of \citet{pang2023frozen} and demonstrated that gradients from different samples with respect to the model weights are more closely aligned in the presence of the frozen LLM block.
The authors quantified this alignment by demonstrating an improved gradient signal-to-noise ratio (GSNR) in the presence of the LLM block.
For a given parameter, GSNR is the ratio between the squared expected value and the variance of its gradient.
A high GSNR is also associated with improved generalization in machine learning models \citep{liu2020understanding, michalkiewicz2023domain}, and is therefore a desirable property.

\citet{bai2025frozen} also showed that this effect is more pronounced in layers closer to the LLM block, and that the representational similarity between the ViT blocks and the LLM block could be indicative of improvements.
Following this observation and taking inspiration from \citet{tiwari2023overcoming}, \citet{bai2025frozen} then propose an auxiliary training objective aimed at removing the additional inference costs incurred by the LLM block.
This auxiliary training objective distills the representations of the frozen-LLM-appended ViT into a vanilla ViT through an intermediate similarity loss \citep{hinton2015distilling}.

The work of \citet{bai2025frozen} thus presents an orthogonal and potentially interesting future direction for our work.
In particular, their auxiliary loss could be combined with \ours by using it as the teacher model for distilling a vanilla ViT, as \ours achieves stronger visual recognition performance than the baseline teacher models utilized by \citet{bai2025frozen}.

\subsection{Comparisons with Large Vision-Language Models}
\label{app-sec:comparison-with-lvlms}
\myparagraph{Large language models for visual tasks.}
Large language models (LLMs) are used together with visual encoders in numerous multimodal architectural settings.
The most common branch of this work involves using LLMs as \textit{textual decoders} \citep{li2022blip, li2023blip, liu2023visual, chen2024expanding, chen2024internvl, alayrac2022flamingo}, preceded by visual encoders.
In these works, visual tokens produced by the encoder are projected directly into the text decoder \citep{li2022blip, liu2023visual} or fused through additional cross-modal layers \citep{alayrac2022flamingo}.

All of the aforementioned works demonstrate that LLMs can process vision-originating data when it is first processed by a separate visual encoder \citep{liu2023visual, li2022blip}, or when the models are trained jointly from scratch using vast amounts of data across multiple stages \citep{diao2024unveiling, wang2025vision, luo2024mono}.
Our work is inspired by the success of these approaches while differing in several key respects.
Specifically, our goal is to effectively leverage LLM transformer blocks and Self-Supervised Learning (SSL) to improve the performance of vision transformers \citep{dosovitskiy2020image}, without relying on language-aligned visual encoders (e.g., CLIP \citep{radford2021learning}) or requiring language inputs.

\myparagraph{Large monolithic vision-language models.}
A related branch of work comprises foundation vision-language models that aim to contain both the vision and language modalities within a large monolithic transformer \citep{diao2024unveiling, diao2025evev2, wang2025vision, luo2024mono, bavishi2023introducing, chen2024single}.
These works differ from \textit{encoder-decoder} \citep{li2022blip, liu2023visual, li2023blip, yu2022coca, wan2024locca} and \textit{two-tower encoder} \citep{radford2021learning, siglip2, zhai2023sigmoid} alternatives in that they enforce varying degrees of intra-block parameter sharing between the transformers for each modality.
For example, while Fuyu \citep{bavishi2023introducing} and EVEv1 \citep{diao2024unveiling} share the majority of their Transformer components, EVEv2 \citep{diao2025evev2} shares only the self-attention block while retaining modality-specific layer normalization (LN) \citep{ba2016layer} and MLP blocks within each transformer.

While monolithic vision-language models share some similarities with our work, they also differ in several key respects.
Specifically, while our goal is to achieve stronger \textit{discriminative} visual performance, these works mainly target generative domains, such as visual question answering (VQA) \citep{goyal2017making} and image captioning \citep{chen2015microsoft}.

In addition, all of these monolithic vision-language model works \citep{diao2024unveiling, diao2025evev2, wang2025vision, luo2024mono, bavishi2023introducing, chen2024single} jointly train both language- and vision-related components on vast amounts of multimodal data across multiple stages and with multiple objectives.
In our work, we instead adapt the LLM block using simple and cost-effective LoRA layers under a unified MAE objective, without requiring language-specific inputs or additional objectives, thereby achieving strong unimodal performance without extensive multimodal training.

\subsection{Comparison with Robust Visual Representation Learning Techniques}
\label{app-sec:robust-vit-comparison}
An existing body of literature focuses on improving the robustness and out-of-distribution generalization of Vision Transformers \citep{ViTinDomain, FAN, RVT, MedViT}.
For example, the Robust Vision Transformer (RVT) \citep{RVT} relies on specialized supervised training techniques, such as Position-Aware Attention Scaling (PAAS) and patch-wise data augmentation, to filter noisy correlations and enhance diversity \citep{RVT}.
Similarly, works such as Fully Attentional Networks (FAN) \citep{FAN} and MedViT \citep{MedViT} introduce domain-specific or related structural modifications to improve resilience against distribution shifts.
In contrast, \ours operates strictly within a self-supervised foundation-model paradigm.
Instead of relying on supervised training augmentations or redesigning the internal structure of standard ViT blocks, \ours achieves enhanced robustness by extracting and injecting high-level semantic priors from a frozen LLM block using a pure Masked Auto-Encoding (MAE) objective.
Because \ours does not conflict with intra-block architectural modifications or supervised data-augmentation strategies, these robust ViT variants present an orthogonal and highly complementary direction.

%% file: TMLR/Supplementary_Material/A_4_Detailed_Training_Configs.tex
\section{Training Details of Experiments}
\label{app-sec:further-training-details}
In this section, we describe the architectural details, hyperparameter settings and other training details that we adhered to throughout this work.

\subsection{Architectural Details}
Throughout our experiments, we utilize the ViT/B as our encoder from \citet{dosovitskiy2020image} with a patch size of $16\text{x}16$, which consists of $12$ Transformer \citep{vaswani2017attention} blocks and has a hidden size of $768$.
In addition, for \ours, we primarily utilize the $32^{nd}$ (i.e the final) Transformer \citep{vaswani2017attention} block of the smallest LLaMA 1 \citep{touvron2023llama} model with 7 billion parameters and a hidden size of $4096$, unless otherwise stated for ablations.
We choose this block of LLaMA 1 following its success in similar works \citep{lai2024residual, pang2023frozen, bai2025frozen} and our empirical observations following our ablations in Section \ref{sec:ablation}.
There are also two additional linear projections \textit{without} any non-linearities or additional activations around the LLaMA 1 block to allow matching the hidden dimensions of the ViT and the LLaMA 1.

During the pretraining stage, for both \ours and our baselines, we additionally employ a lightweight Transformer \citep{vaswani2017attention} decoder, which consists of $8$ blocks and has a hidden size of $512$.
The design of both the ViT/B encoder and the lightweight decoder closely mirror the original MAE design with no changes with the exception of the LLaMA 1 block and the linear projections around it.

For LoRA-related hyperparameters, we performed our experiments with a rank of $r=16$ throughout our work, following the common usage of this parameter in the literature \citep{hu2022lora}.
Similarly, following \citet{hu2022lora}, we set the alpha parameter the same as our rank parameter, \textit{i.e.} $r=\alpha=16$.
We also provide additional ablations on the rank hyperparameter and $\alpha$ hyperparameters in Section \ref{app-sec:additional-ablations-exp}.
Finally, we did not adjust a particular learning rate scheduling mechanism for LoRA layers, and they directly followed the learning rate schedule of the final ViT block.

\subsection{Self-supervised Pretraining}
\label{app-sec:ssl-details}
For all of our self-supervised pretraining experiments, we directly adhere to all of the settings presented in the original MAE work \citep{he2022masked}, while training both our models and the baselines for $800$ epochs.

Namely, this corresponds to having a batch size of $4096$, base learning rate of $1.5e\text{-}04$, with cosine annealing scheduling \citep{loshchilov2016sgdr}.
In addition, we used the AdamW optimizer \citep{kingma2014adam, loshchilov2017decoupled} with $\beta_1 = 0.90$ and $\beta_2 = 0.95$ \citep{chen2020generative}, coupled with $40$ warm-up epochs \citep{goyal2017accurate} and a weight decay of $0.05$.
Finally, we also apply a random resized crop augmentation, utilized a random masking ratio of $75\%$ for masking the encoder inputs, and a normalized pixel version of mean squared error (MSE) between the reconstructed and the ground truth images as the objective.
\subsection{End-to-end Finetuning for Classification}
\label{app-sec:cls-e2e-details}
Analogously with Section \ref{app-sec:ssl-details}, we directly adhere to all of the settings presented in the original MAE work \citep{he2022masked}.
Namely, this corresponds to having a batch size of $1024$, learning rate of $1.e\text{-}03$, with cosine annealing scheduling \citep{loshchilov2016sgdr}.
In addition, we used the AdamW optimizer \citep{kingma2014adam, loshchilov2017decoupled} with $\beta_1 = 0.90$ and $\beta_2 = 0.999$ \citep{chen2020generative}, coupled with $5$ warm-up epochs \citep{goyal2017accurate} and a weight decay of $0.05$.
Differing from the pretraining stage, here we have a layer-wise learning rate decay value of $0.75$ \citep{bao2021beit, clark2020electra}, label smoothing of $0.1$ \citep{szegedy2016rethinking} and a drop path rate of $0.1$ \citep{huang2016deep}.
Finally, we also applied \textit{mixup} \citep{zhang2017mixup} with $0.8$, \textit{cutmix} \cite{yun2019cutmix} with $1.0$, and Randaugment with ($9, 0.5$) \citep{cubuk2020randaugment}.

Notably, we utilize average pooling setting instead of relying on the [CLS] token for performing classification.
We do so, following the official MAE Github repository's \footnote{\url{https://github.com/facebookresearch/mae/tree/main}} report of potential instabilities in the loss values \footnote{\url{https://github.com/facebookresearch/mae/blob/main/FINETUNE.md}} when the [CLS] token was used with Pytorch \citep{paszke2019pytorch}.

\subsection{Training for Fine-grained Visual Recognition}
\label{app-sec:det-e2e-details}
Our fine-grained visual recognition experiments mostly follow from the ViTDet framework \citep{li2022exploring}, which is a competitive fine-grained visual recognition framework achieving competitive results with plain ViT backbones \citep{dosovitskiy2020image} with respect to previously-stronger hierarchical counterparts, such as the Swin Transformer \citep{liu2021swin}.
ViTDet framework involves taking an MAE pretrained plain ViT backbone, a following simple feature pyramid structure \citet{lin2017feature} and a Mask R-CNN \citep{he2017mask} as the final detection/segmentation head.
Notably, achieving competitive fine-grained visual recognition results is very hard with supervised-only ViT backbones, with neither of Imagenet-1K nor Imagenet-22K supervised-pretrained ViT/B models achieving better results than a randomly initialized ViT/B, further highlighting the necessity of self-supervised pretraining for achieving strong fine-grained visual recognition.

Finally, the entire model, with the notable exception of the LLM block that we always keep frozen and merely adapt through the LoRA layers, including the ViT/\ours backbones, is trained jointly on the COCO training set \citep{lin2014mscoco}, with a batch size of $64$, a learning rate of $1.5e\text{-}04$, weight decay of $0.1$, drop path rate of $0.1$ and for $100$ epochs.
For both our baselines and \ours, we directly adhere to the settings of \citet{li2022exploring}, and do not change any hyperparameters.
We implemented our \ours/B ViTDet and benchmarked both \ours and our baselines on the mmdetection library \citep{chen2019mmdetection}.
\subsection{Computational Resources}
\label{app-sec:computational-resources}
For all of the aforementioned experiments, we ran our experiments on $32$ NVIDIA A100 GPUs.
For MAE pretraining described in Section \ref{app-sec:ssl-details}, the baseline experiments take approximately $30$ hours with experiments with \ours taking slightly longer around $35$ hours.
For both the end-to-end finetuning for classification and the fine-grained visual recognition training experiments, both \ours and the baseline experiments took approximately $24$ hours.

%% file: TMLR/Supplementary_Material/A_5_Detailed_Datasets.tex
\section{Details of the Used Datasets}
\label{app-sec:dataset-descriptions}
In this section, we provide the details of the datasets we used for our experiments and other quantitative analyses, while clarifying the exact splits and settings we report our results on.
\paragraph{Imagenet-1K.} Imagenet-1K \citep{deng2009imagenet} consists of $1.2M$ training and $50K$ validation images, belonging to $1000$ different classes.
Following the conventional approach \citep{he2016deep, dosovitskiy2020image}, we used the resized ($224^2$) images for both training and evaluation.
We performed the MAE pretraining exclusively on Imagenet-1K training set, for all of our classification and fine-grained visual recognition experiments, following our baselines \citep{he2022masked, li2022exploring}.
\paragraph{Imagenet-9.} Imagenet-9 \citep{xiao2020noise} consists of images of the $9$ super-classes from the original Imagenet-1K validation set \citep{deng2009imagenet}, and aims to measure the background over-reliance of deep learning models in an evaluation-only setting.
In particular, Imagenet-9 has contains numerous splits, such as the \textit{original}, \textit{mixed random}, \textit{mixed same}, and \textit{mixed next}.
The first of these splits, \textit{original} consists of the unaltered images belonging to the $9$ super-classes, with their original backgrounds.
On the other hand, \textit{mixed random, mixed same}, and \textit{mixed next} consist of images with altered backgrounds.
For \textit{mixed random}, the background of each image is replaced with the background of another image from a random super-class, for \textit{mixed next}, the background of each image is replaced with the background of another image from the next super-class ordered with respect to their numerical IDs, and for \textit{mixed same} the background of each image is replaced with the background of another image from the same super-class.

In Imagenet-9 \citep{xiao2020noise}, while it is desirable to obtain high performance on the clean \textit{original} set, it is crucial to obtain high performance on the splits with altered backgrounds for demonstrating the robustness of the models, thereby achieving a smaller \textit{background accuracy gap}.
Finally, for our evaluations, we utilized the \textit{original} split for benchmarking the clean accuracy of the models in our work, while comparing it to the accuracies in \textit{mixed random} and \textit{mixed same} splits for measuring the background over-reliance of models.
\paragraph{Imagenet-Segmentation.} Imagenet-Segmentation \citep{gao2022large} consists of the images and associated high-quality segmentation masks of the original Imagenet-1K \citep{deng2009imagenet} images.
It has $3$ splits of different sizes, Imagenet-Segmentation-50 as the $50$ class subset with $752$ validation images, Imagenet-Segmentation-300 as the $300$ class subset with $4K$ validation images, and Imagenet-Segmentation-919 as the $919$ class subset with $12K$ validation images.
Notably, the largest $919$ split does not contain the images of non-segmentable $81$ classes from the original Imagenet-1K splits \citep{gao2022large}.
We re-purpose this dataset in the same format as \citet{pang2023frozen}, though including additional results and visualizations on the more challenging Imagenet-Segmentation-300 and Imagenet-Segmentation-919 instead of limiting the analyses to the limited Imagenet-Segmentation-50 split as in \citet{pang2023frozen}.
\paragraph{Imagenet-C.} Imagenet-C \citep{hendrycks2019benchmarking} benchmark is an evaluation-only benchmark consists of synthetically corrupted images belonging to the Imagenet-1K validation \citep{deng2009imagenet} set.
In particular, there are $15$ benchmark corruptions, namely $4$ noise corruptions (\textit{gaussian noise, shot noise, impulse noise}), $4$ weather-related corruptions (\textit{snow, frost, fog, brightness}), $4$ blurring corruptions (\textit{defocus, glass, motion, zoom}) and $3$ digital corruptions (\textit{contrast, elastic transform, pixelate}).
Furthermore, there are $4$ additional corruptions, namely \textit{gaussian blur, spatter, saturate, speckle noise}, bringing the total to $19$.
For each of these corruptions, there are $5$ severity levels, with higher number indicating tougher corruptions.
In our experiments, we report the average results on all of the aforementioned corruptions with all of their severities for a more comprehensive evaluation.
\paragraph{Imagenet-A.} Imagenet-A \citep{hendrycks2021natural} is an adversarially-designed benchmark consisting of images from Imagenet-1K validation set, where the majority of the Imagenet-1K-trained classifiers fail.
Notably, it has $200$ super-classes instead of the full $1000$ classes of the Imagenet-1K benchmark, where the super-classes were explicitly constructed in a way that confusing them would be beyond a simple confusion of similar classes.
\paragraph{Imagenet-SK.} Imagenet-SK \citep{wang2019learning} consists of $50K$ images of sketches of Imagenet-1K classes, $50$ for each of the $1000$ classes of the Imagenet-1K validation set.
Notably, images of Imagenet-SK are \textit{black and white} sketches, posing a challenge due to their lack of texture and color information.
\paragraph{Imagenet-V2.} Imagenet-V2 \citep{recht2019imagenet} is a benchmark proposed to measure the broader generalization capabilities of Imagenet-1K-trained models.
It also has samples for the same $1000$ classes of the Imagenet-1K, though with specifically curated examples where the majority of the Imagenet-1K-trained classifiers tend to fail.
Among its different variants, we utilized the \textit{matched frequency} version, as it is proposed to be the default setting in \citet{recht2019imagenet}.
\paragraph{Imagenet-R.} Imagenet-R \citep{hendrycks2021many} is a $30K$ image domain-generalization benchmark for Imagenet-1K-trained classifiers.
It contains \textit{``renditions''} of images belonging to Imagenet-1K classes, in the form of images of sculptures or paintings, with drastically different textures, and other often-helpful image-level statistics.

\paragraph{Imagenet-E.} Imagenet-E \citep{li2023imagenete} is an evaluation benchmark designed to systematically measure the robustness of image classifiers against specific, fine-grained visual attribute shifts.
It consists of images where foreground objects and backgrounds have been explicitly manipulated to isolate different types of vulnerabilities.
The benchmark includes splits for background perturbations (such as varying texture complexities, random backgrounds, and adversarial backgrounds) as well as geometric and spatial transformations (including varying object sizes, random object positions, and random directions).
By isolating these attributes, Imagenet-E provides a detailed assessment of whether a model relies on robust semantic object features rather than contextual or spatial biases.

\paragraph{MS COCO.} MS COCO \citep{lin2014mscoco} is an object detection and instance segmentation benchmark for benchmarking fine-grained visual recognition capabilities of deep learning models.
Among its variants, we train both \ours/B with ViTDet \citep{li2022exploring} and ViT/B with ViTDet \citep{li2022exploring} models on COCO2017 training set and report our results on the COCO2017 validation set, following the common practice \citep{li2022exploring, carion2020end, he2017mask}.

%% file: TMLR/Supplementary_Material/A_6_Limitations.tex
\section{Limitations}
\label{app-sec:limitations}
Even though \ours benefits from the combined powers of self-supervised learning with MAE and the LoRA-adapted pretrained LLM representations for discriminative computer vision tasks, it also inherits the drawbacks of these works.
First, the two-stage training nature of our framework (MAE pretraining followed by end-to-end finetuning) involves substantial computational cost, albeit still being comparable to standard self-supervised learning pipelines. 
Crucially, we strictly utilize parameter-efficient fine-tuning (PEFT) via LoRA to adapt the LLM component. Because we update a negligible fraction of the LLM parameters ($\sim0.3\%$), the \textit{training} overhead, both in terms of GPU memory (VRAM) requirements and backward-pass compute is not drastically higher than that of a standard MAE ViT baseline, keeping the pre-training phase highly accessible.

Conversely, the addition of the large LLM block inevitably introduces a heavier computational footprint during \textit{forward-pass inference}, which may limit its usage in downstream tasks requiring strict real-time processing.
Specifically, the inclusion of the LLM block increases the inference complexity from approximately 17.6G FLOPs (standard ViT/B) to 59.0G FLOPs for \ours/B, with a corresponding peak memory footprint increase from 0.4GB to 1.3GB.

Regardless, as highlighted by \ours's significant performance and robustness improvements over models of similar parameter sizes (\textit{i.e.,} the randomly-initialized LLM block in Section \ref{sec:experiments} and the trainable LLM block ablations in Appendix \ref{app-sec:trainable-llm-block}), \ours's combination of pretrained LLM representations, the MAE objective, and LoRA adaptation proves highly desirable for domains where robustness and out-of-distribution generalization are prioritized over strict real-time inference latency.

%% file: main.bib
@article{ViTinDomain,
author = {Alijani, Shadi and Fayyad, Jamil and Najjaran, Homayoun},
title = {Vision transformers in domain adaptation and domain generalization: a study of robustness},
year = {2024},
issue_date = {Oct 2024},
publisher = {Springer-Verlag},
address = {Berlin, Heidelberg},
volume = {36},
number = {29},
issn = {0941-0643},
url = {https://doi.org/10.1007/s00521-024-10353-5},
doi = {10.1007/s00521-024-10353-5},
journal = {Neural Comput. Appl.},
month = aug,
pages = {17979–18007},
numpages = {29}
}

@article{MedViT,
title = {MedViT: A robust vision transformer for generalized medical image classification},
journal = {Computers in Biology and Medicine},
volume = {157},
pages = {106791},
year = {2023},
issn = {0010-4825},
doi = {https://doi.org/10.1016/j.compbiomed.2023.106791},
url = {https://www.sciencedirect.com/science/article/pii/S0010482523002561},
author = {Omid Nejati Manzari and Hamid Ahmadabadi and Hossein Kashiani and Shahriar B. Shokouhi and Ahmad Ayatollahi},
}

@inproceedings{FAN,
  title={Understanding the robustness in vision transformers},
  author={Zhou, Daquan and Yu, Zhiding and Xie, Enze and Xiao, Chaowei and Anandkumar, Animashree and Feng, Jiashi and Alvarez, Jose M},
  booktitle={International conference on machine learning},
  pages={27378--27394},
  year={2022},
  organization={PMLR}
}

@inproceedings{RVT,
  title={Towards robust vision transformer},
  author={Mao, Xiaofeng and Qi, Gege and Chen, Yuefeng and Li, Xiaodan and Duan, Ranjie and Ye, Shaokai and He, Yuan and Xue, Hui},
  booktitle={Proceedings of the IEEE/CVF conference on Computer Vision and Pattern Recognition},
  pages={12042--12051},
  year={2022}
}

@article{raghu2021vision,
  title={Do vision transformers see like convolutional neural networks?},
  author={Raghu, Maithra and Unterthiner, Thomas and Kornblith, Simon and Zhang, Chiyuan and Dosovitskiy, Alexey},
  journal={Advances in neural information processing systems},
  volume={34},
  pages={12116--12128},
  year={2021}
}

@inproceedings{li2023imagenete,
  title={Imagenet-e: Benchmarking neural network robustness via attribute editing},
  author={Li, Xiaodan and Chen, Yuefeng and Zhu, Yao and Wang, Shuhui and Zhang, Rong and Xue, Hui},
  booktitle={Proceedings of the IEEE/CVF Conference on Computer Vision and Pattern Recognition},
  pages={20371--20381},
  year={2023}
}

@article{gao2022large,
  title={Large-scale unsupervised semantic segmentation},
  author={Gao, Shanghua and Li, Zhong-Yu and Yang, Ming-Hsuan and Cheng, Ming-Ming and Han, Junwei and Torr, Philip},
  journal={IEEE transactions on pattern analysis and machine intelligence},
  volume={45},
  number={6},
  pages={7457--7476},
  year={2022},
  publisher={IEEE}
}

@article{pont20172017,
  title={The 2017 davis challenge on video object segmentation},
  author={Pont-Tuset, Jordi and Perazzi, Federico and Caelles, Sergi and Arbelaez, Pablo and Sorkine-Hornung, Alex and Van Gool, Luc},
  journal={arXiv preprint arXiv:1704.00675},
  year={2017}
}

@inproceedings{lai2024residual,
  title={Residual-based language models are free boosters for biomedical imaging tasks},
  author={Lai, Zhixin and Wu, Jing and Chen, Suiyao and Zhou, Yucheng and Hovakimyan, Naira},
  booktitle={Proceedings of the IEEE/CVF Conference on Computer Vision and Pattern Recognition},
  pages={5086--5096},
  year={2024}
}

@inproceedings{li2022blip,
  title={Blip: Bootstrapping language-image pre-training for unified vision-language understanding and generation},
  author={Li, Junnan and Li, Dongxu and Xiong, Caiming and Hoi, Steven},
  booktitle={International conference on machine learning},
  pages={12888--12900},
  year={2022},
  organization={PMLR}
}

@article{liu2023visual,
  title={Visual instruction tuning},
  author={Liu, Haotian and Li, Chunyuan and Wu, Qingyang and Lee, Yong Jae},
  journal={Advances in neural information processing systems},
  volume={36},
  pages={34892--34916},
  year={2023}
}

@article{pang2023frozen,
    author={Pang, Ziqi and Xie, Ziyang and Man, Yunze and Wang, Yu-Xiong},
  title={Frozen transformers in language models are effective visual encoder layers},
  journal={arXiv preprint arXiv:2310.12973},
  year={2023}
}

@article{touvron2023llama,
  title={Llama: Open and efficient foundation language models},
  author={Touvron, Hugo and Lavril, Thibaut and Izacard, Gautier and Martinet, Xavier and Lachaux, Marie-Anne and Lacroix, Timoth{\'e}e and Rozi{\`e}re, Baptiste and Goyal, Naman and Hambro, Eric and Azhar, Faisal and others},
  journal={arXiv preprint arXiv:2302.13971},
  year={2023}
}

@article{alayrac2022flamingo,
  title={Flamingo: a visual language model for few-shot learning},
  author={Alayrac, Jean-Baptiste and Donahue, Jeff and Luc, Pauline and Miech, Antoine and Barr, Iain and Hasson, Yana and Lenc, Karel and Mensch, Arthur and Millican, Katherine and Reynolds, Malcolm and others},
  journal={Advances in neural information processing systems},
  volume={35},
  pages={23716--23736},
  year={2022}
}

@inproceedings{li2023blip,
  title={Blip-2: Bootstrapping language-image pre-training with frozen image encoders and large language models},
  author={Li, Junnan and Li, Dongxu and Savarese, Silvio and Hoi, Steven},
  booktitle={International conference on machine learning},
  pages={19730--19742},
  year={2023},
  organization={PMLR}
}

@article{dosovitskiy2020image,
  title={An image is worth 16x16 words: Transformers for image recognition at scale},
  author={Dosovitskiy, Alexey and Beyer, Lucas and Kolesnikov, Alexander and Weissenborn, Dirk and Zhai, Xiaohua and Unterthiner, Thomas and Dehghani, Mostafa and Minderer, Matthias and Heigold, Georg and Gelly, Sylvain and others},
  journal={arXiv preprint arXiv:2010.11929},
  year={2020}
}

@article{vaswani2017attention,
  title={Attention is all you need},
  author={Vaswani, Ashish and Shazeer, Noam and Parmar, Niki and Uszkoreit, Jakob and Jones, Llion and Gomez, Aidan N and Kaiser, {\L}ukasz and Polosukhin, Illia},
  journal={Advances in neural information processing systems},
  volume={30},
  year={2017}
}

@article{grattafiori2024llama,
  title={The llama 3 herd of models},
  author={Grattafiori, Aaron and Dubey, Abhimanyu and Jauhri, Abhinav and Pandey, Abhinav and Kadian, Abhishek and Al-Dahle, Ahmad and Letman, Aiesha and Mathur, Akhil and Schelten, Alan and Vaughan, Alex and others},
  journal={arXiv preprint arXiv:2407.21783},
  year={2024}
}

@article{touvron2023llama2,
  title={Llama 2: Open foundation and fine-tuned chat models},
  author={Touvron, Hugo and Martin, Louis and Stone, Kevin and Albert, Peter and Almahairi, Amjad and Babaei, Yasmine and Bashlykov, Nikolay and Batra, Soumya and Bhargava, Prajjwal and Bhosale, Shruti and others},
  journal={arXiv preprint arXiv:2307.09288},
  year={2023}
}

@article{team2024gemma,
  title={Gemma 2: Improving open language models at a practical size},
  author={Team, Gemma and Riviere, Morgane and Pathak, Shreya and Sessa, Pier Giuseppe and Hardin, Cassidy and Bhupatiraju, Surya and Hussenot, L{\'e}onard and Mesnard, Thomas and Shahriari, Bobak and Ram{\'e}, Alexandre and others},
  journal={arXiv preprint arXiv:2408.00118},
  year={2024}
}

@inproceedings{he2022masked,
  title={Masked autoencoders are scalable vision learners},
  author={He, Kaiming and Chen, Xinlei and Xie, Saining and Li, Yanghao and Doll{\'a}r, Piotr and Girshick, Ross},
  booktitle={Proceedings of the IEEE/CVF conference on computer vision and pattern recognition},
  pages={16000--16009},
  year={2022}
}

@article{oquab2023dinov2,
  title={Dinov2: Learning robust visual features without supervision},
  author={Oquab, Maxime and Darcet, Timoth{\'e}e and Moutakanni, Th{\'e}o and Vo, Huy and Szafraniec, Marc and Khalidov, Vasil and Fernandez, Pierre and Haziza, Daniel and Massa, Francisco and El-Nouby, Alaaeldin and others},
  journal={arXiv preprint arXiv:2304.07193},
  year={2023}
}

@article{wang2025vision,
  title={Vision as LoRA},
  author={Wang, Han and Ye, Yongjie and Li, Bingru and Nie, Yuxiang and Lu, Jinghui and Tang, Jingqun and Wang, Yanjie and Huang, Can},
  journal={arXiv preprint arXiv:2503.20680},
  year={2025}
}

@article{diao2025evev2,
  title={EVEv2: Improved Baselines for Encoder-Free Vision-Language Models},
  author={Diao, Haiwen and Li, Xiaotong and Cui, Yufeng and Wang, Yueze and Deng, Haoge and Pan, Ting and Wang, Wenxuan and Lu, Huchuan and Wang, Xinlong},
  journal={arXiv preprint arXiv:2502.06788},
  year={2025}
}

@article{diao2024unveiling,
  title={Unveiling encoder-free vision-language models},
  author={Diao, Haiwen and Cui, Yufeng and Li, Xiaotong and Wang, Yueze and Lu, Huchuan and Wang, Xinlong},
  journal={arXiv preprint arXiv:2406.11832},
  year={2024}
}

@article{luo2024mono,
  title={Mono-internvl: Pushing the boundaries of monolithic multimodal large language models with endogenous visual pre-training},
  author={Luo, Gen and Yang, Xue and Dou, Wenhan and Wang, Zhaokai and Liu, Jiawen and Dai, Jifeng and Qiao, Yu and Zhu, Xizhou},
  journal={arXiv preprint arXiv:2410.08202},
  year={2024}
}

@article{hu2022lora,
  title={Lora: Low-rank adaptation of large language models.},
  author={Hu, Edward J and Shen, Yelong and Wallis, Phillip and Allen-Zhu, Zeyuan and Li, Yuanzhi and Wang, Shean and Wang, Lu and Chen, Weizhu and others},
  journal={ICLR},
  volume={1},
  number={2},
  pages={3},
  year={2022}
}

@article{su2024roformer,
  title={Roformer: Enhanced transformer with rotary position embedding},
  author={Su, Jianlin and Ahmed, Murtadha and Lu, Yu and Pan, Shengfeng and Bo, Wen and Liu, Yunfeng},
  journal={Neurocomputing},
  volume={568},
  pages={127063},
  year={2024},
  publisher={Elsevier}
}

@inproceedings{caron2021emerging,
  title={Emerging properties in self-supervised vision transformers},
  author={Caron, Mathilde and Touvron, Hugo and Misra, Ishan and J{\'e}gou, Herv{\'e} and Mairal, Julien and Bojanowski, Piotr and Joulin, Armand},
  booktitle={Proceedings of the IEEE/CVF international conference on computer vision},
  pages={9650--9660},
  year={2021}
}

@inproceedings{devlin2019bert,
  title={Bert: Pre-training of deep bidirectional transformers for language understanding},
  author={Devlin, Jacob and Chang, Ming-Wei and Lee, Kenton and Toutanova, Kristina},
  booktitle={Proceedings of the 2019 conference of the North American chapter of the association for computational linguistics: human language technologies, volume 1 (long and short papers)},
  pages={4171--4186},
  year={2019}
}

@article{grill2020bootstrap,
  title={Bootstrap your own latent-a new approach to self-supervised learning},
  author={Grill, Jean-Bastien and Strub, Florian and Altch{\'e}, Florent and Tallec, Corentin and Richemond, Pierre and Buchatskaya, Elena and Doersch, Carl and Avila Pires, Bernardo and Guo, Zhaohan and Gheshlaghi Azar, Mohammad and others},
  journal={Advances in neural information processing systems},
  volume={33},
  pages={21271--21284},
  year={2020}
}

@inproceedings{chen2020simple,
  title={A simple framework for contrastive learning of visual representations},
  author={Chen, Ting and Kornblith, Simon and Norouzi, Mohammad and Hinton, Geoffrey},
  booktitle={International conference on machine learning},
  pages={1597--1607},
  year={2020},
  organization={PmLR}
}

@inproceedings{he2020momentum,
  title={Momentum contrast for unsupervised visual representation learning},
  author={He, Kaiming and Fan, Haoqi and Wu, Yuxin and Xie, Saining and Girshick, Ross},
  booktitle={Proceedings of the IEEE/CVF conference on computer vision and pattern recognition},
  pages={9729--9738},
  year={2020}
}

@article{vincent2010stacked,
  title={Stacked denoising autoencoders: Learning useful representations in a deep network with a local denoising criterion.},
  author={Vincent, Pascal and Larochelle, Hugo and Lajoie, Isabelle and Bengio, Yoshua and Manzagol, Pierre-Antoine and Bottou, L{\'e}on},
  journal={Journal of machine learning research},
  volume={11},
  number={12},
  year={2010}
}

@inproceedings{li2022exploring,
  title={Exploring plain vision transformer backbones for object detection},
  author={Li, Yanghao and Mao, Hanzi and Girshick, Ross and He, Kaiming},
  booktitle={European conference on computer vision},
  pages={280--296},
  year={2022},
  organization={Springer}
}

@inproceedings{chen2020generative,
  title={Generative pretraining from pixels},
  author={Chen, Mark and Radford, Alec and Child, Rewon and Wu, Jeffrey and Jun, Heewoo and Luan, David and Sutskever, Ilya},
  booktitle={International conference on machine learning},
  pages={1691--1703},
  year={2020},
  organization={PMLR}
}

@article{bao2021beit,
  title={Beit: Bert pre-training of image transformers},
  author={Bao, Hangbo and Dong, Li and Piao, Songhao and Wei, Furu},
  journal={arXiv preprint arXiv:2106.08254},
  year={2021}
}

@article{zhou2021ibot,
  title={ibot: Image bert pre-training with online tokenizer},
  author={Zhou, Jinghao and Wei, Chen and Wang, Huiyu and Shen, Wei and Xie, Cihang and Yuille, Alan and Kong, Tao},
  journal={arXiv preprint arXiv:2111.07832},
  year={2021}
}

@article{chen2024expanding,
  title={Expanding performance boundaries of open-source multimodal models with model, data, and test-time scaling},
  author={Chen, Zhe and Wang, Weiyun and Cao, Yue and Liu, Yangzhou and Gao, Zhangwei and Cui, Erfei and Zhu, Jinguo and Ye, Shenglong and Tian, Hao and Liu, Zhaoyang and others},
  journal={arXiv preprint arXiv:2412.05271},
  year={2024}
}

@inproceedings{chen2024internvl,
  title={Internvl: Scaling up vision foundation models and aligning for generic visual-linguistic tasks},
  author={Chen, Zhe and Wu, Jiannan and Wang, Wenhai and Su, Weijie and Chen, Guo and Xing, Sen and Zhong, Muyan and Zhang, Qinglong and Zhu, Xizhou and Lu, Lewei and others},
  booktitle={Proceedings of the IEEE/CVF conference on computer vision and pattern recognition},
  pages={24185--24198},
  year={2024}
}

@inproceedings{bai2025frozen,
  title={Frozen Language Models Are Gradient Coherence Rectifiers in Vision Transformers},
  author={Bai, Lichen and Xiong, Zixuan and Lin, Hai and Xu, Guangwei and Xie, Xiangjin and Guo, Ruijie and Kang, Zhanhui and Zheng, Hai-Tao and Kim, Hong-Gee},
  booktitle={Proceedings of the AAAI Conference on Artificial Intelligence},
  volume={39},
  pages={1817--1825},
  year={2025}
}

@inproceedings{radford2021learning,
  title={Learning transferable visual models from natural language supervision},
  author={Radford, Alec and Kim, Jong Wook and Hallacy, Chris and Ramesh, Aditya and Goh, Gabriel and Agarwal, Sandhini and Sastry, Girish and Askell, Amanda and Mishkin, Pamela and Clark, Jack and others},
  booktitle={International conference on machine learning},
  pages={8748--8763},
  year={2021},
  organization={PmLR}
}

@inproceedings{deng2009imagenet,
  title={Imagenet: A large-scale hierarchical image database},
  author={Deng, Jia and Dong, Wei and Socher, Richard and Li, Li-Jia and Li, Kai and Fei-Fei, Li},
  booktitle={2009 IEEE conference on computer vision and pattern recognition},
  pages={248--255},
  year={2009},
  organization={Ieee}
}

@inproceedings{hendrycks2021natural,
  title={Natural adversarial examples},
  author={Hendrycks, Dan and Zhao, Kevin and Basart, Steven and Steinhardt, Jacob and Song, Dawn},
  booktitle={Proceedings of the IEEE/CVF conference on computer vision and pattern recognition},
  pages={15262--15271},
  year={2021}
}

@inproceedings{wang2019learning,
        title={Learning Robust Global Representations by Penalizing Local Predictive Power},
        author={Wang, Haohan and Ge, Songwei and Lipton, Zachary and Xing, Eric P},
        booktitle={Advances in Neural Information Processing Systems},
        pages={10506--10518},
        year={2019}
}

@inproceedings{recht2019imagenet,
  title={Do imagenet classifiers generalize to imagenet?},
  author={Recht, Benjamin and Roelofs, Rebecca and Schmidt, Ludwig and Shankar, Vaishaal},
  booktitle={International conference on machine learning},
  pages={5389--5400},
  year={2019},
  organization={PMLR}
}

@inproceedings{hendrycks2021many,
  title={The many faces of robustness: A critical analysis of out-of-distribution generalization},
  author={Hendrycks, Dan and Basart, Steven and Mu, Norman and Kadavath, Saurav and Wang, Frank and Dorundo, Evan and Desai, Rahul and Zhu, Tyler and Parajuli, Samyak and Guo, Mike and others},
  booktitle={Proceedings of the IEEE/CVF international conference on computer vision},
  pages={8340--8349},
  year={2021}
}

@article{hendrycks2019benchmarking,
  title={Benchmarking neural network robustness to common corruptions and perturbations},
  author={Hendrycks, Dan and Dietterich, Thomas},
  journal={arXiv preprint arXiv:1903.12261},
  year={2019}
}

@article{xiao2020noise,
  title={Noise or signal: The role of image backgrounds in object recognition},
  author={Xiao, Kai and Engstrom, Logan and Ilyas, Andrew and Madry, Aleksander},
  journal={arXiv preprint arXiv:2006.09994},
  year={2020}
}

@inproceedings{lin2014mscoco,
  title={Microsoft coco: Common objects in context},
  author={Lin, Tsung-Yi and Maire, Michael and Belongie, Serge and Hays, James and Perona, Pietro and Ramanan, Deva and Doll{\'a}r, Piotr and Zitnick, C Lawrence},
  booktitle={Computer vision--ECCV 2014: 13th European conference, zurich, Switzerland, September 6-12, 2014, proceedings, part v 13},
  pages={740--755},
  year={2014},
  organization={Springer}
}

@inproceedings{lin2017feature,
  title={Feature pyramid networks for object detection},
  author={Lin, Tsung-Yi and Doll{\'a}r, Piotr and Girshick, Ross and He, Kaiming and Hariharan, Bharath and Belongie, Serge},
  booktitle={Proceedings of the IEEE conference on computer vision and pattern recognition},
  pages={2117--2125},
  year={2017}
}

@inproceedings{he2017mask,
  title={Mask r-cnn},
  author={He, Kaiming and Gkioxari, Georgia and Doll{\'a}r, Piotr and Girshick, Ross},
  booktitle={Proceedings of the IEEE international conference on computer vision},
  pages={2961--2969},
  year={2017}
}

@inproceedings{guo2023robustifying,
  title={Robustifying token attention for vision transformers},
  author={Guo, Yong and Stutz, David and Schiele, Bernt},
  booktitle={Proceedings of the IEEE/CVF International Conference on Computer Vision},
  pages={17557--17568},
  year={2023}
}

@inproceedings{carion2020end,
  title={End-to-end object detection with transformers},
  author={Carion, Nicolas and Massa, Francisco and Synnaeve, Gabriel and Usunier, Nicolas and Kirillov, Alexander and Zagoruyko, Sergey},
  booktitle={European conference on computer vision},
  pages={213--229},
  year={2020},
  organization={Springer}
}

@article{siglip2,
  title={Siglip 2: Multilingual vision-language encoders with improved semantic understanding, localization, and dense features},
  author={Tschannen, Michael and Gritsenko, Alexey and Wang, Xiao and Naeem, Muhammad Ferjad and Alabdulmohsin, Ibrahim and Parthasarathy, Nikhil and Evans, Talfan and Beyer, Lucas and Xia, Ye and Mustafa, Basil and others},
  journal={arXiv preprint arXiv:2502.14786},
  year={2025}
}

@inproceedings{brown2020language,
 author = {Brown, Tom and Mann, Benjamin and Ryder, Nick and Subbiah, Melanie and Kaplan, Jared D and Dhariwal, Prafulla and Neelakantan, Arvind and Shyam, Pranav and Sastry, Girish and Askell, Amanda and Agarwal, Sandhini and Herbert-Voss, Ariel and Krueger, Gretchen and Henighan, Tom and Child, Rewon and Ramesh, Aditya and Ziegler, Daniel and Wu, Jeffrey and Winter, Clemens and Hesse, Chris and Chen, Mark and Sigler, Eric and Litwin, Mateusz and Gray, Scott and Chess, Benjamin and Clark, Jack and Berner, Christopher and McCandlish, Sam and Radford, Alec and Sutskever, Ilya and Amodei, Dario},
 booktitle = {Advances in Neural Information Processing Systems},
 editor = {H. Larochelle and M. Ranzato and R. Hadsell and M.F. Balcan and H. Lin},
 pages = {1877--1901},
 publisher = {Curran Associates, Inc.},
 title = {Language Models are Few-Shot Learners},
 url = {https://proceedings.neurips.cc/paper_files/paper/2020/file/1457c0d6bfcb4967418bfb8ac142f64a-Paper.pdf},
 volume = {33},
 year = {2020}
}

@inproceedings{zhai2023stabilizing,
  title={Stabilizing transformer training by preventing attention entropy collapse},
  author={Zhai, Shuangfei and Likhomanenko, Tatiana and Littwin, Etai and Busbridge, Dan and Ramapuram, Jason and Zhang, Yizhe and Gu, Jiatao and Susskind, Joshua M},
  booktitle={International Conference on Machine Learning},
  pages={40770--40803},
  year={2023},
  organization={PMLR}
}

@article{liang2022mind,
  title={Mind the gap: Understanding the modality gap in multi-modal contrastive representation learning},
  author={Liang, Victor Weixin and Zhang, Yuhui and Kwon, Yongchan and Yeung, Serena and Zou, James Y},
  journal={Advances in Neural Information Processing Systems},
  volume={35},
  pages={17612--17625},
  year={2022}
}

@inproceedings{naeem2024silc,
  title={Silc: Improving vision language pretraining with self-distillation},
  author={Naeem, Muhammad Ferjad and Xian, Yongqin and Zhai, Xiaohua and Hoyer, Lukas and Van Gool, Luc and Tombari, Federico},
  booktitle={European Conference on Computer Vision},
  pages={38--55},
  year={2024},
  organization={Springer}
}

@article{zhang2024attention,
  title={Attention Entropy is a Key Factor: An Analysis of Parallel Context Encoding with Full-attention-based Pre-trained Language Models},
  author={Zhang, Zhisong and Wang, Yan and Huang, Xinting and Fang, Tianqing and Zhang, Hongming and Deng, Chenlong and Li, Shuaiyi and Yu, Dong},
  journal={arXiv preprint arXiv:2412.16545},
  year={2024}
}

@article{loshchilov2016sgdr,
  title={Sgdr: Stochastic gradient descent with warm restarts},
  author={Loshchilov, Ilya and Hutter, Frank},
  journal={arXiv preprint arXiv:1608.03983},
  year={2016}
}

@article{kingma2014adam,
  title={Adam: A method for stochastic optimization},
  author={Kingma, Diederik P},
  journal={arXiv preprint arXiv:1412.6980},
  year={2014}
}

@article{loshchilov2017decoupled,
  title={Decoupled weight decay regularization},
  author={Loshchilov, Ilya and Hutter, Frank},
  journal={arXiv preprint arXiv:1711.05101},
  year={2017}
}

@article{goyal2017accurate,
  title={Accurate, large minibatch sgd: Training imagenet in 1 hour},
  author={Goyal, Priya and Doll{\'a}r, Piotr and Girshick, Ross and Noordhuis, Pieter and Wesolowski, Lukasz and Kyrola, Aapo and Tulloch, Andrew and Jia, Yangqing and He, Kaiming},
  journal={arXiv preprint arXiv:1706.02677},
  year={2017}
}

@article{paszke2019pytorch,
  title={Pytorch: An imperative style, high-performance deep learning library},
  author={Paszke, A},
  journal={arXiv preprint arXiv:1912.01703},
  year={2019}
}

@article{clark2020electra,
  title={Electra: Pre-training text encoders as discriminators rather than generators},
  author={Clark, Kevin and Luong, Minh-Thang and Le, Quoc V and Manning, Christopher D},
  journal={arXiv preprint arXiv:2003.10555},
  year={2020}
}

@inproceedings{szegedy2016rethinking,
  title={Rethinking the inception architecture for computer vision},
  author={Szegedy, Christian and Vanhoucke, Vincent and Ioffe, Sergey and Shlens, Jon and Wojna, Zbigniew},
  booktitle={Proceedings of the IEEE conference on computer vision and pattern recognition},
  pages={2818--2826},
  year={2016}
}

@inproceedings{huang2016deep,
  title={Deep networks with stochastic depth},
  author={Huang, Gao and Sun, Yu and Liu, Zhuang and Sedra, Daniel and Weinberger, Kilian Q},
  booktitle={Computer Vision--ECCV 2016: 14th European Conference, Amsterdam, The Netherlands, October 11--14, 2016, Proceedings, Part IV 14},
  pages={646--661},
  year={2016},
  organization={Springer}
}

@inproceedings{cubuk2020randaugment,
  title={Randaugment: Practical automated data augmentation with a reduced search space},
  author={Cubuk, Ekin D and Zoph, Barret and Shlens, Jonathon and Le, Quoc V},
  booktitle={Proceedings of the IEEE/CVF conference on computer vision and pattern recognition workshops},
  pages={702--703},
  year={2020}
}

@article{zhang2017mixup,
  title={mixup: Beyond empirical risk minimization},
  author={Zhang, Hongyi and Cisse, Moustapha and Dauphin, Yann N and Lopez-Paz, David},
  journal={arXiv preprint arXiv:1710.09412},
  year={2017}
}

@inproceedings{yun2019cutmix,
  title={Cutmix: Regularization strategy to train strong classifiers with localizable features},
  author={Yun, Sangdoo and Han, Dongyoon and Oh, Seong Joon and Chun, Sanghyuk and Choe, Junsuk and Yoo, Youngjoon},
  booktitle={Proceedings of the IEEE/CVF international conference on computer vision},
  pages={6023--6032},
  year={2019}
}

@inproceedings{touvron2021training,
  title={Training data-efficient image transformers \& distillation through attention},
  author={Touvron, Hugo and Cord, Matthieu and Douze, Matthijs and Massa, Francisco and Sablayrolles, Alexandre and J{\'e}gou, Herv{\'e}},
  booktitle={International conference on machine learning},
  pages={10347--10357},
  year={2021},
  organization={PMLR}
}

@inproceedings{liu2021swin,
  title={Swin transformer: Hierarchical vision transformer using shifted windows},
  author={Liu, Ze and Lin, Yutong and Cao, Yue and Hu, Han and Wei, Yixuan and Zhang, Zheng and Lin, Stephen and Guo, Baining},
  booktitle={Proceedings of the IEEE/CVF international conference on computer vision},
  pages={10012--10022},
  year={2021}
}

@article{chen2019mmdetection,
  title={MMDetection: Open mmlab detection toolbox and benchmark},
  author={Chen, Kai and Wang, Jiaqi and Pang, Jiangmiao and Cao, Yuhang and Xiong, Yu and Li, Xiaoxiao and Sun, Shuyang and Feng, Wansen and Liu, Ziwei and Xu, Jiarui and others},
  journal={arXiv preprint arXiv:1906.07155},
  year={2019}
}

@inproceedings{tiwari2023overcoming,
  title={Overcoming simplicity bias in deep networks using a feature sieve},
  author={Tiwari, Rishabh and Shenoy, Pradeep},
  booktitle={International Conference on Machine Learning},
  pages={34330--34343},
  year={2023},
  organization={PMLR}
}

@article{hinton2015distilling,
  title={Distilling the knowledge in a neural network},
  author={Hinton, Geoffrey and Vinyals, Oriol and Dean, Jeff},
  journal={arXiv preprint arXiv:1503.02531},
  year={2015}
}

@inproceedings{zhai2023sigmoid,
  title={Sigmoid loss for language image pre-training},
  author={Zhai, Xiaohua and Mustafa, Basil and Kolesnikov, Alexander and Beyer, Lucas},
  booktitle={Proceedings of the IEEE/CVF international conference on computer vision},
  pages={11975--11986},
  year={2023}
}

@article{yu2022coca,
  title={Coca: Contrastive captioners are image-text foundation models},
  author={Yu, Jiahui and Wang, Zirui and Vasudevan, Vijay and Yeung, Legg and Seyedhosseini, Mojtaba and Wu, Yonghui},
  journal={arXiv preprint arXiv:2205.01917},
  year={2022}
}

@article{wan2024locca,
  title={Locca: Visual pretraining with location-aware captioners},
  author={Wan, Bo and Tschannen, Michael and Xian, Yongqin and Pavetic, Filip and Alabdulmohsin, Ibrahim M and Wang, Xiao and Susano Pinto, Andr{\'e} and Steiner, Andreas and Beyer, Lucas and Zhai, Xiaohua},
  journal={Advances in Neural Information Processing Systems},
  volume={37},
  pages={116355--116387},
  year={2024}
}

@article{bavishi2023introducing,
  title={Introducing our multimodal models, 2023},
  author={Bavishi, Rohan and Elsen, Erich and Hawthorne, Curtis and Nye, Maxwell and Odena, Augustus and Somani, Arushi and Tas{\i}rlar, Sagnak},
  journal={URL https://www. adept. ai/blog/fuyu-8b},
  volume={2},
  year={2023}
}

@article{chen2024single,
  title={A Single Transformer for Scalable Vision-Language Modeling},
  author={Chen, Yangyi and Wang, Xingyao and Peng, Hao and Ji, Heng},
  journal={arXiv preprint arXiv:2407.06438},
  year={2024}
}

@article{ba2016layer,
  title={Layer normalization},
  author={Ba, Jimmy Lei and Kiros, Jamie Ryan and Hinton, Geoffrey E},
  journal={arXiv preprint arXiv:1607.06450},
  year={2016}
}

@inproceedings{goyal2017making,
  title={Making the v in vqa matter: Elevating the role of image understanding in visual question answering},
  author={Goyal, Yash and Khot, Tejas and Summers-Stay, Douglas and Batra, Dhruv and Parikh, Devi},
  booktitle={Proceedings of the IEEE conference on computer vision and pattern recognition},
  pages={6904--6913},
  year={2017}
}

@article{chen2015microsoft,
  title={Microsoft coco captions: Data collection and evaluation server},
  author={Chen, Xinlei and Fang, Hao and Lin, Tsung-Yi and Vedantam, Ramakrishna and Gupta, Saurabh and Doll{\'a}r, Piotr and Zitnick, C Lawrence},
  journal={arXiv preprint arXiv:1504.00325},
  year={2015}
}

@inproceedings{he2016deep,
  title={Deep residual learning for image recognition},
  author={He, Kaiming and Zhang, Xiangyu and Ren, Shaoqing and Sun, Jian},
  booktitle={Proceedings of the IEEE conference on computer vision and pattern recognition},
  pages={770--778},
  year={2016}
}

@article{liu2020understanding,
  title={Understanding why neural networks generalize well through gsnr of parameters},
  author={Liu, Jinlong and Jiang, Guoqing and Bai, Yunzhi and Chen, Ting and Wang, Huayan},
  journal={arXiv preprint arXiv:2001.07384},
  year={2020}
}

@inproceedings{michalkiewicz2023domain,
  title={Domain generalization guided by gradient signal to noise ratio of parameters},
  author={Michalkiewicz, Mateusz and Faraki, Masoud and Yu, Xiang and Chandraker, Manmohan and Baktashmotlagh, Mahsa},
  booktitle={Proceedings of the IEEE/CVF International Conference on Computer Vision},
  pages={6177--6188},
  year={2023}
}
